\documentclass[review,lettersize,journal]{IEEEtran}
\usepackage{xspace}
\usepackage{graphicx} 
\usepackage{amsmath} 
\usepackage{amsfonts}  
\usepackage{subcaption}
\usepackage{xcolor}
\usepackage{dsfont}
\usepackage{url}
\usepackage[normalem]{ulem}
\usepackage{booktabs} 
\usepackage{amsmath,amssymb,amsfonts,pifont}
\usepackage{algorithm}%
\usepackage{algpseudocode}%
\usepackage{graphicx,textcomp}
\usepackage{soul,longtable,lscape,mathabx} %
\usepackage{xcolor,hyperref}
\usepackage{xr-hyper}
\newif\ifanonymous

\newcommand{\F}{\ensuremath{\mathcal{F}}\xspace}  
\newcommand{\A}{\ensuremath{\mathcal{A}}\xspace}  
\newcommand{\B}{\ensuremath{\mathcal{B}}\xspace}  
\newcommand{\E}{\ensuremath{\mathcal{E}}\xspace}  
\newcommand{\U}{\ensuremath{\mathcal{U}}\xspace}
\newcommand{\MS}{\ensuremath{\mathcal{S}}\xspace}  
\newcommand{\fopt}{\ensuremath{f^{\text{opt}}}\xspace} 
\newcommand{\fbest}{\ensuremath{f^{\text{best}}}\xspace} 
\newcommand{\efbest}{\ensuremath{f^{\text{best}}}\xspace} 
\newcommand{\af}{\ensuremath{\phi\xspace}} 
\newcommand{\eaf}{\ensuremath{\widehat{\phi}\xspace}} 
\newcommand{\perf}{\ensuremath{\mathcal{J}\xspace}} 
\newcommand{\size}[1]{\ensuremath{\left|#1\right|}\xspace} 
\newcommand{\UB}{\ensuremath{\mathbb{U}}\xspace} 
\newcommand{\LB}{\ensuremath{\mathbb{B}}\xspace} 
\newcommand{\multi}{\ensuremath{\mu}\xspace} 

\newcommand{\pair}[2]{\ensuremath{\left( #1, #2 \right)}}
\newcommand{\penalise}{\ensuremath{\pi}\xspace}

\AtBeginDocument{%
  }
\newcommand{\new}[1]{#1}

\begin{document}
\title{How Sequential Algorithm Portfolios can Benefit Black Box Optimization}

\ifanonymous
  \author{Anonymous Authors}
\else
  \author{Catalin-Viorel Dinu,
        Diederick Vermetten,
        Carola Doerr
        \thanks{Catalin-Viorel Dinu (Catalin-Viorel.Dinu@lip6.fr), Diederick Vermetten (Diederick.Vermetten@lip6.fr) and Carola Doerr (Carola.Doerr@lip6.fr) are with Sorbonne Université, CNRS, LIP6 in Paris, France}}
\fi

\maketitle
\begin{abstract}

\new{In typical black-box optimization applications, the available computational budget is often allocated to a single algorithm, usually selected based on user preference or expert advice. When little is known about the problem, however, committing the entire budget to one solver can be risky. In this work, we conduct a post-hoc investigation of the potential benefits of distributing a fixed budget across multiple algorithms using historical performance data. To support this investigation, we introduce a simple yet efficient approach for constructing sequential portfolios and evaluate them using performance measures derived from empirical attainment functions (EAFs). This EAF-based perspective provides a flexible way to capture performance across multiple target levels and utility preferences. The benefits of using multiple algorithms arise from two complementary effects: differences in algorithm performance across diverse problems and variance reduction among stochastic solvers. Importantly, these advantages can be realized even when parallel evaluations are unavailable.}

\new{To assess the potential of sequential algorithm portfolios, we apply our approach to the COCO data archive, using performance data from more than 200 algorithms evaluated on the BBOB test suite. The resulting portfolios consistently outperform single-algorithm baselines, achieving relative performance gains of more than 14\%. Moreover, these gains persist across a range of utility-weighted performance objectives, including settings in which all weight is placed on the most demanding target. This result indicates that the observed improvements are not confined to average-case performance.}


\end{abstract}
\section{Introduction}
In the context of black-box optimization, a vast collection of algorithms has been developed over the years, each tailored to address specific characteristics or challenges inherent to certain classes of problems. This naturally leads to complementarity among algorithms: different optimizers are better suited for different problem types. The best choice depends on the structure of the instance, the available evaluation budget, and the performance metric being optimized.

This diversity gives rise to the classical question: which algorithm should be used when solving a specific optimization problem (or a set of problems)? This is the core of the well-studied \emph{algorithm selection problem}~\cite{journals/ac/Rice76}. Modern solutions typically fall into two categories:
\begin{itemize}
    \item General-purpose recommendation system, based on ``a~priori'' knowledge such as variable types or a budget constraint~\cite{meunier2021black}.
    \item Feature-based selection methods, where part of the evaluation budget is used to compute problem-specific features (commonly through Exploratory Landscape Analysis (ELA)~\cite{mersmann2011exploratory}, or other sampling-based approaches~\cite{cenikj2026survey}), and the algorithm is chosen based on these.
\end{itemize}
However, both approaches come with risks. The extracted features might not fully capture the problem's true structure, or the a priori knowledge may be insufficient or misleading. As such, selecting a single algorithm based on such information may lead to sub-optimal performance.

\new{In this work, we do not propose a deployable alternative to algorithm selection. Instead, we conduct a purely post-hoc analysis to quantify the extent to which distributing the available budget across multiple algorithms could improve performance when the most suitable solver is not known in advance. Establishing the magnitude of this potential benefit is an important prerequisite for the later development of practical portfolio-selection methods for unseen problems. To make this analysis possible at scale, we also introduce a simple and computationally efficient method for constructing sequential portfolios from a large archive of historical algorithm-performance data}

\new{Distributing evaluations across several algorithms may reduce reliance on any single solver and thereby improve robustness. At the same time, it may limit the resources available to an algorithm that would have performed particularly well if allowed to continue. Whether a portfolio is advantageous therefore depends on the algorithms involved, their performance distributions, the problem considered, and the available evaluation budget~\cite{lindauer2015sequential}}

\new{A prominent instance of this idea is the parallel portfolio, in which multiple algorithms are executed concurrently. Such portfolios have demonstrated strong empirical performance in domains including SAT solving~\cite{xu2008satzilla}. These findings do not imply that portfolios universally outperform their individual components, but rather suggest that combining complementary algorithms can be beneficial under suitable conditions.}

\new{Parallel portfolios can reduce runtime variability by spreading risk across multiple algorithms, thereby limiting the impact of an unfavorable run by any single solver. This effect is not universal, however, and depends on factors such as the algorithms’ runtime distributions and their dependence. At the same time, parallel execution can limit the opportunity to capitalize on algorithms that perform exceptionally well when allowed to continue running. This trade-off is especially pronounced in settings where performance improves substantially with longer uninterrupted runs, as parallel portfolios may terminate promising runs earlier than a single-algorithm strategy would.}  However, rather than treating algorithm selection and parallel portfolios as opposing strategies, we propose to combine their strengths. While this has been tackled in online settings, where budget is allocated based on comparisons of progress between algorithms in the portfolio, e.g., in bet-and-run strategies~\cite{friedrich2017generic}, we focus here on a purely offline strategy without any needed interaction between algorithms. Specifically, we ask: \textbf{How can a fixed evaluation budget be optimally divided among a set of algorithms?}

By allowing heterogeneous budget allocations, we are no longer restricted to uniform or parallel execution. This formulation provides a continuum; from allocating the entire budget to a single best-performing algorithm, to finely splitting the budget across several complementary algorithms. This flexibility enables the construction of portfolios that adapt to both performance variation and algorithmic complementarity, making better use of the available evaluation budget.


\new{In this paper, we address the budget allocation problem from a feature-free, data-driven perspective. Rather than relying on exploratory features or prior knowledge about the problem landscape, we propose a method that determines how to distribute a fixed evaluation budget across multiple algorithms based solely on existing performance data. This approach is conceptually similar to selecting a single best solver (SBS), but extends it by enabling heterogeneous budget allocation across multiple independent algorithm runs. Portfolio performance is evaluated using empirical attainment functions (EAFs), which capture performance jointly across evaluation budgets and target precisions. We further introduce utility-weighted variants of this performance measure, allowing the portfolio construction to reflect different preferences over solution quality, ranging from uniform importance across targets to concentrating all importance on the most demanding target. Our results show that exploiting complementary algorithm behavior can lead to substantial improvements over the SBS. Additional experiments also show that these gains cannot be explained by simple equal budget splitting across independent restarts alone.}

\new{In the following sections, we introduce the portfolio construction problem (Section~\ref{sec:algorithm_portfolio}), define the EAF-based performance metric and baselines (Sections~\ref{sec:portfolio_performance}--\ref{sec:baselines}), and present the main experimental results (Section~\ref{sec:main_results}). We then describe the greedy portfolio construction method (Section~\ref{sec:methods}) and investigate several extensions and analysis settings in Section~\ref{sec:extra}, including utility-weighted portfolio objectives (Section~\ref{sec:utility}), incremental portfolio construction (Section~\ref{sec:incremental}), and the robustness of the method under randomly sampled algorithm sets (Section~\ref{sec:random}).}

\textbf{Reproducibility:} 
Our code  to reproduce results is available at~\cite{zenodo}. For most parts, we use freely accessible data from the COCO data repository, available at \cite{coco_data_archive}.

\section{Algorithm Portfolios}
\label{sec:algorithm_portfolio}
\subsection{Related Work}

\new{
Algorithm portfolios combine multiple algorithms or configurations to exploit complementary behavior and improve robustness across problem instances. They have traditionally been studied in parallel settings, where several solvers are executed simultaneously. Such portfolios may be assembled from a set of strong algorithms or constructed automatically to promote complementarity among their components~\cite{xu2008satzilla,xu2010hydra,lindauer2017automatic}. Similar ideas have been applied across domains, including SAT, planning, and continuous optimization~\cite{gomes2001algorithm,tang2021few,schede2025method}.
}

\new{
Portfolio methods can also operate sequentially, with the available budget divided among algorithms according to a fixed or automatically generated schedule~\cite{kadioglu2011algorithm,lindauer2016empirical}. Existing approaches represent such portfolios as sequences of algorithm-runtime pairs and allocate computation to components based on their expected contribution to portfolio performance~\cite{streeter2007combining,seipp2015automatic}. Alternatively, portfolio creation can be seen as sequential selection of runs of its component algorithms~\cite{HE2019559, schapermeier2025greedy}, which are thus adapted based on the success of the previous component of the portfolio.  
}

\new{
Sequential execution does not necessarily imply chaining. In scheduling-based portfolios, components may be executed one after another while remaining independent. This differs from approaches in which the output or internal state of one algorithm is passed to the next. A similar distinction appears in adversarial robustness evaluation: serial attack policies use the output of one attack to initialize the following attack, whereas attack ensembles execute their components individually and combine only their outcomes~\cite{liu2023reliable}.
}

\new{
The portfolios studied here are sequential only in their execution order. Each run is initialized independently, no solutions or internal states are transferred, and later runs do not depend on earlier outcomes. Repeated occurrences of the same algorithm therefore correspond to independent restarts rather than continued execution. Performance gains arise from distributing the budget across multiple runs and exploiting stochastic variation or complementary algorithm behavior.
}

\new{
We evaluate these portfolios using the empirical attainment function (EAF), which describes the probability of attaining a given solution quality within a given budget. Unlike a single final-quality or fixed-cutoff measure, the EAF captures anytime behavior across the joint time-quality space. It also provides flexibility in defining the desired portfolio behavior: different regions can be emphasized to prioritize rapid attainment of acceptable solutions, reliable attainment of demanding targets, or balanced performance across a range of budgets and quality levels. The EAF therefore serves both as an evaluation tool and as a basis for customizing the portfolio objective.
}

\subsection{Problem Formulation}\label{sec:problem_formulation}

\new{We formulate portfolio construction as a \textbf{Budget Allocation} problem. Given a fixed total evaluation budget T, the goal is to determine which algorithm-budget pairs should be included in the portfolio, and with what multiplicity, such that the resulting portfolio performance is maximized while \emph{the total allocated budget does not exceed T}. Each portfolio component is executed independently, using an independent random seed and without sharing solutions, states, or other information between runs. This leads to the following constrained optimization problem:}

\begin{align*}
    \max_{\MS \in \Pi} \quad & \perf(\F,\MS) \\
    \text{s.t.} \quad & \sum_{\pair{\alpha}{b} \in \MS} b \leq T
\end{align*}
Where $\perf$ is the performance measure to be maximized, $\mathcal{F}$ denotes the set of optimization problems under consideration, and $\Pi$ represents the set of all feasible portfolios. A portfolio is defined as a multiset $\MS = \{\pair{\alpha}{b} \mid \alpha \in \A, b \in \B\}$ consisting of pairs of algorithms and their allocated budgets, where $\mathcal{A}$ is the set of admissible algorithms and $\mathcal{B}$ is the set of admissible budget allocations.

Note that we use a multi-set representation here, since we can utilize multiple runs of the same algorithm to benefit from the variance reduction this brings. 

We observe that the structure of our portfolio optimization problem closely resembles the Unbounded Knapsack Problem (UKP) \cite{hu2008unbounded}, where the goal is to select items (in our case, algorithm–budget pairs) under a total cost constraint in order to maximize some reward. However, a key distinction in our setting is that the performance function $\perf$ can be highly non-linear, due to the interaction between multiple algorithm-budget pairs across multiple functions.

\subsection{Portfolio Performance}\label{sec:portfolio_performance}

\begin{figure}[tbp]
    \centering
    
    \begin{subfigure}[b]{0.49\columnwidth}
        \centering
        \includegraphics[width=\textwidth]{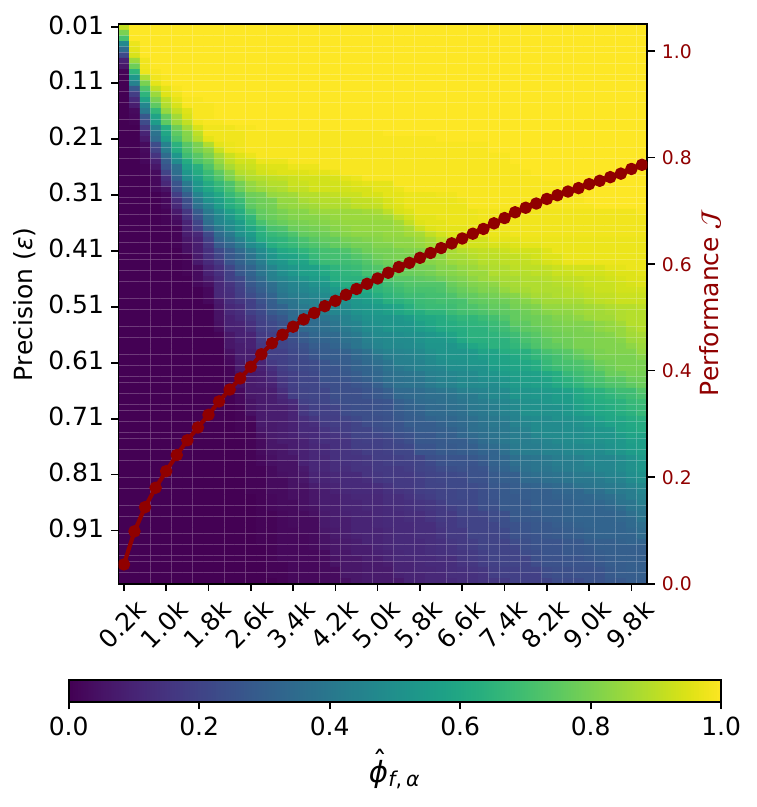}
        \caption{BBOB $F_6$\\CMAES (nevergrad).\\Steady improvement with budget}
        \label{fig:eaf_example_1}
    \end{subfigure}
    \hfill
    \begin{subfigure}[b]{0.49\columnwidth}
        \centering
        \includegraphics[width=\textwidth]{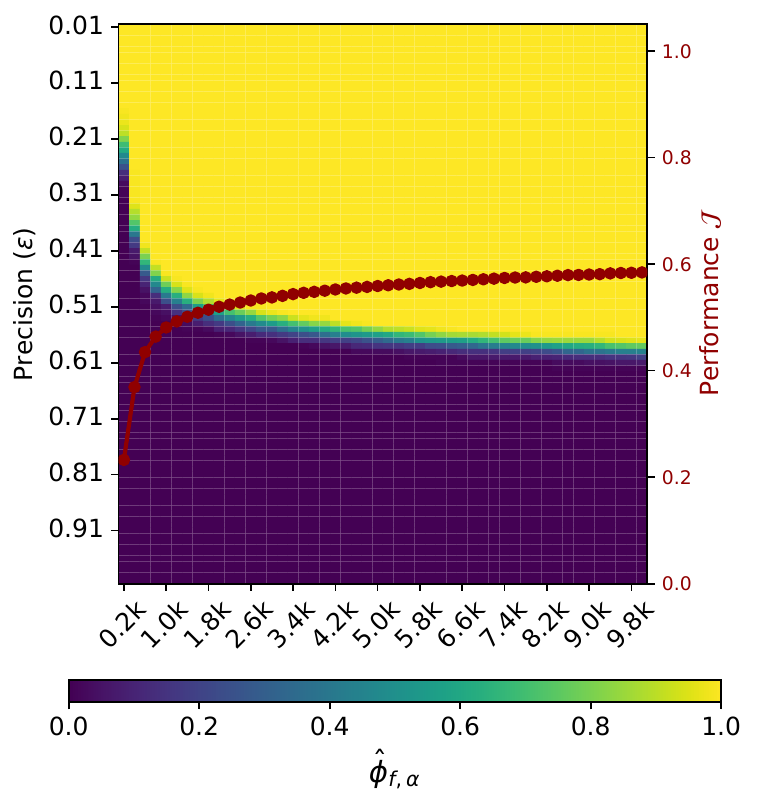}
        \caption{BBOB $F_{14}$\\CMAES (nevergrad)\\Early performance plateau}
        \label{fig:eaf_example_2}
    \end{subfigure}

    \caption{Empirical attainment plots for BBOB functions using CMA-ES from nevergrad. The horizontal axis shows function evaluations ($\B$), and the vertical axis represents normalized precisions ($\E$). The red curve represent the performance of a portfolio consisting only of one algorithm-budget pair.}
    \label{fig:eaf_example}
\end{figure}

As performance criterion, we make use of the empirical attainment function \cite{grunert2001inferential, lopez2024using} to represent an algorithm's anytime performance. For any function $f\in\F$, let $\fopt$ denote the global minimum of $f$. For any algorithm $\alpha\in\A$ and any function $f\in\F$, define $\fbest_{\alpha}(b)$ as the best (smallest) function value that is seen by $\alpha$ on $f$ within a budget of $b$ function evaluations.
Given this, we can define the \textit{attainment function} for target precision $\varepsilon>0$ as: 
\begin{equation}
    \af_{f, \alpha}(b, \varepsilon) = \mathbb{P}\left[\fbest_{\alpha}(b) - \fopt \leq \varepsilon \right].
\end{equation}

We can estimate the attainment function by executing the algorithm $R$ times and recording the values $\fbest_{\alpha,i}(b)$, $i=1,\ldots,R$ of best seen values in these $R$ runs after the first $b$ function evaluations. 
\begin{equation}
     \eaf_{f,\alpha}(b,\varepsilon)
  = \frac{1}{R}\sum_{i=1}^{R}
    \mathds{1}_{\left(\,\efbest_{\alpha,i}(b) - \fopt \le \varepsilon\,\right)}.
\end{equation}
where $\mathds{1}_{\xi}$ denotes the indicator function which returns $1$ if event  $\xi$ happens and $0$ otherwise. 

We discretize both the budget and precision levels to obtain a matrix of empirical values (see Figure~\ref{fig:eaf_example} for examples). Specifically, we define $\B = \{b_i\}$ as the set of discrete budget levels (typically chosen to be evenly spaced) and $\E = \{\varepsilon_j\}$ as the set of target precision thresholds. For visualization purposes, we normalize the precision axis to lie within the interval $[0,1]$, where a value of $0$ corresponds to failure to attain even the coarsest precision level (i.e., $\max \E$), and a value of $1$ corresponds to successful attainment of the finest precision level (i.e., $\min \E$).
\begin{figure*}[htbp]
    \centering
    \includegraphics[width=\linewidth]{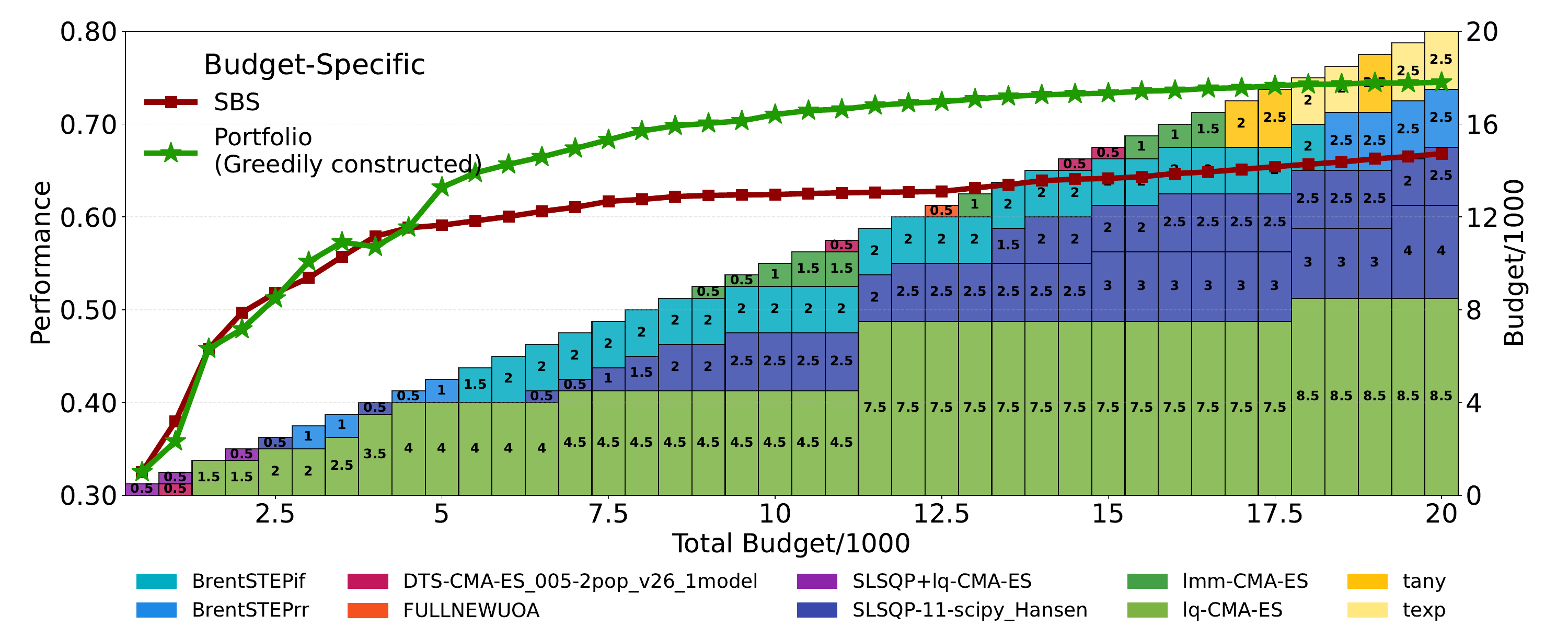}
    \caption{Performance comparison between the portfolio and baseline algorithm over increasing total budgets, using a subset of COCO’s algorithm set. Background elements represent the composition of the resulting portfolio. We observe that performance gains become more prominent as the total budget increases, and the selected portfolios diversify accordingly.}
    \label{fig:portofolios_total_budget}
\end{figure*}
In this setting, we are essentially evaluating the empirical cumulative distribution function (ECDF) over a range of precision thresholds. A natural way to summarize performance is to compute the average empirical attainment across all precision levels in $\E$. Thus, for a single algorithm–budget pair, we define its performance:
\begin{equation}
    \perf(f,\alpha) = \frac{1}{\size{\E}}
    \sum_{\varepsilon \in \E} \eaf_{f,\alpha}(b, \varepsilon).
\end{equation}
as the mean attainment over all $\varepsilon \in \E$. An illustration of how this metric evolves with increasing budget is shown in Figure~\ref{fig:eaf_example}, where we observe contrasting algorithm behaviors: steady performance improvements in Figure~\ref{fig:eaf_example_1}, and early saturation with performance plateaus in Figure~\ref{fig:eaf_example_2}.

We now generalize this formulation to the portfolio setting, allowing the portfolio $\MS$ to contain multiple algorithm–budget pairs. At the same time, we extend the performance metric to account for a set of functions $\F$ rather than a single instance.

In this context, we are interested in the expected probability of success: the probability that at least one configuration in the portfolio reaches a given precision threshold $\varepsilon$ on a function $f \in \F$. This is equivalent to one minus the joint probability of failure across all configurations in $\MS$.

To aggregate this performance across the benchmark, we compute the average over all functions in $\mathcal{F}$ and all precision thresholds in $\mathcal{E}$. The resulting portfolio performance metric is defined as:
\begin{equation}\label{eq:performance}
\begin{aligned}
    &\perf(\F,\MS) = \\
    \frac{1}{\size{\F}\size{\E}} \sum_{f \in \F}
    \sum_{\varepsilon \in \E}  &\left( 1 - \prod_{\pair{\alpha}{b} \in \MS}  
    \left(1- \eaf_{f,\alpha}(b, \varepsilon)\right)\right).
\end{aligned}
\end{equation}

\subsection{Performance Baselines}\label{sec:baselines} 
Before introducing our results, we begin by analyzing the structure of the problem and identifying a simple yet informative baseline. For any task, a natural baseline is obtained by selecting a single algorithm that performs best on average across all functions in $\F$, and assigning the entire available budget $T$ to it (commonly referred to as the Single Best Solver, SBS, in the algorithm selection context).
\begin{equation}
    \LB = \max_{\alpha \in \A }  \perf(\F,\{\pair{\alpha}{T}\}). 
\end{equation}
This baseline allows us to evaluate whether more complex portfolios (with multiple smaller-budget configurations) bring a significant advantage over simply committing to the globally strongest single algorithm.

The upper bound on portfolio performance arises from the observation that, under our performance metric, algorithm–budget pairs can be reused. If a given precision level (target) has a non-zero empirical success probability for at least one algorithm, then by allocating more runs (i.e., more budget) to that pair, the chance of reaching that target increases. In the limit, with infinitely many independent repetitions, the success probability for such a target approaches~1. 

This leads to an idealized situation where, for each function and each precision level that is attainable (i.e., where the empirical attainment function is non-zero), we assume perfect success. In other words, the upper bound assumes that for each function we can select the algorithm that eventually reaches all attainable precisions with probability 1, provided an unconstrained and sufficiently large budget,

\begin{equation}
    \UB = \frac{1}{\size{\F} \size{\E}} 
    \sum_{f \in \F} 
    \max_{\alpha \in \A} 
    \left( 
    \sum_{\varepsilon \in \E} 
    \mathds{1}_{ \left(\eaf_{f,\alpha}(b, \varepsilon)> 0\right)} 
    \right).
\end{equation}

\new{While this construction does not necessarily create a tight bound that is achievable, it rather provides a useful metric for performance quantification. Specifically, we can use it to normalize} how much of the theoretically achievable performance a portfolio captures. This leads to the metric we use: relative improvement over the baseline, defined as:
\begin{equation}
\frac{\perf(\F, \MS) - \LB}{\UB - \LB}.
\end{equation}

\begin{figure*}[htbp]
    \centering
    \includegraphics[width=\linewidth]{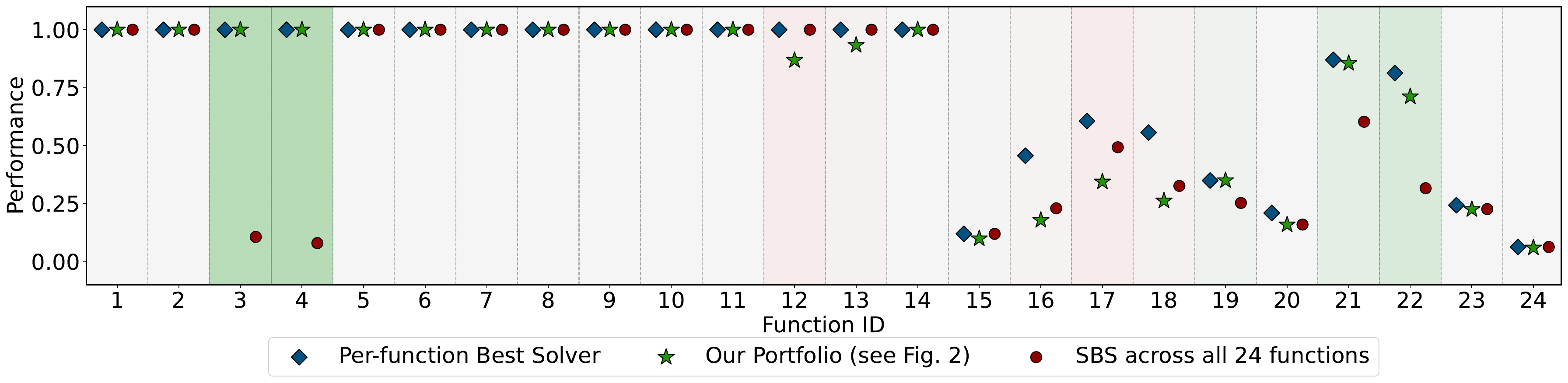}
    \caption{The plot shows: (blue-diamond, leftmost icon in each column) the best-performing algorithm for each function; (green-star, middle) the performance of the constructed portfolio for a total budget of 10,000 function evaluations; and (red-circle, right) the single best algorithm across all functions. As a visual aid, the background color of each column indicates how much better (green) or worse (red) the portfolio-based solver is compared to the single-best solver. Although the portfolio may not outperform the per-function best algorithm, it achieves a higher overall performance by leveraging complementary strengths across the function set.}
    \label{fig:functions_best_coco}
\end{figure*}
\begin{figure*}[htbp]
    \centering
    \includegraphics[width=\linewidth]{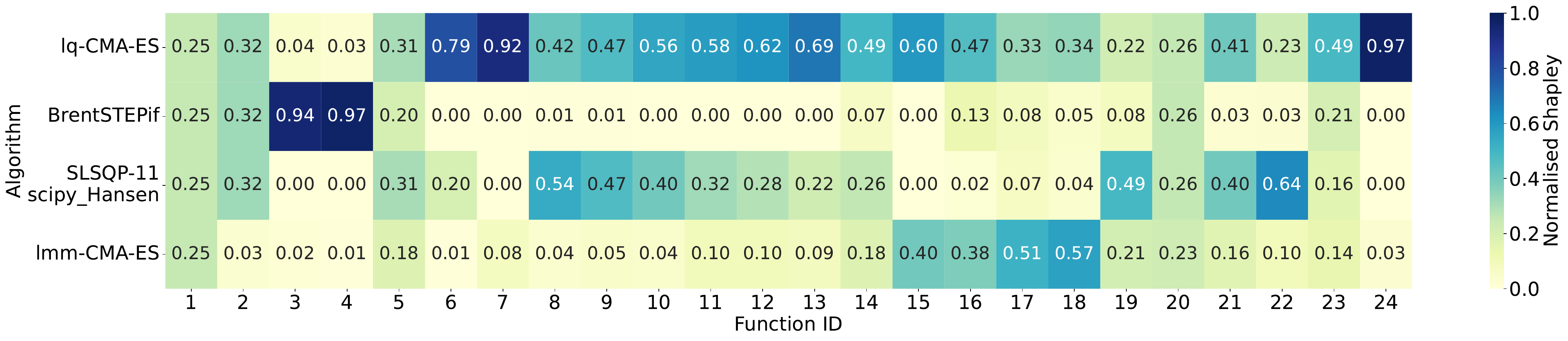}
    \caption{Normalized Shapley values indicating the contribution of each algorithm to each individual function in the portfolio for 10,000 function evaluations. Higher values suggest greater importance of the corresponding algorithm in improving performance for a given function.}
    \label{fig:shaply}
\end{figure*}
\section{Main Results}\label{sec:main_results}
To truly grasp how portfolio construction can contribute to the development of well-structured algorithm sets that outperform on average individual methods, we consider the full set of 24 functions from the BBOB suite~\cite{finck2010real}. These functions span a range of characteristics: separable ($F_1–F_5$), low to moderate conditioning ($F_6–F_9$), highly conditioned unimodal ($F_{10}–F_{14}$), multimodal with a clear global structure ($F_{15}–F_{19}$), and multimodal with weak global structure ($F_{20}–F_{24}$). This diverse benchmark plays a central role in evaluating gradient-free optimization algorithms and has been widely adopted in the optimization community.

To support consistent benchmarking, several archiving initiatives have emerged to collect performance data from a wide range of optimization algorithms. One of the most prominent resources is the COCO archive~\cite{hansen2021coco}, which includes performance data for over 250 algorithms contributed by various researchers and institutions. In our study, we restrict our attention to algorithms evaluated on 10-dimensional BBOB functions and for which 20 independent runs were recorded per function (this being the standard protocol). After applying these filters, we obtain a curated set of 234 algorithms.

Due to the computational and memory overhead of handling such a large-scale dataset, we reduce it to a more manageable subset for meaningful experimentation. Specifically, for each BBOB function, we select the algorithm that achieves the highest average empirical attainment function (EAF) across all budgets and precision levels. This results in a curated set of 20 top-performing algorithms: BIPOP-saACM-k, BrentSTEPrr, DE-BFGS, DTS-CMA-ES\_005-2pop\_v26\_1model, FULLNEWUOA, SLSQP+lq-CMA-ES, BrentSTEPif, SLSQP-11-scipy\_Hansen, lmm-CMA-ES, lq-CMA-ES, tany, texp, NEWUOA, BFGS-M-17, BIRMIN, xNES, BFGS-P-StPt, R-DE-10e5, MLSL, and oMads-Neg.

This reduced selection strikes a balance between computational feasibility and diverse algorithmic coverage, enabling more efficient evaluation of portfolio-based methods while maintaining relevance to real-world algorithm selection challenges. 

We investigate this scenario under varying total budget levels $T_k=500k$, $k \in [1..40]$. For each total budget $T_k$, we define the corresponding intermediary budget set as $\B_k = \{500j\}_{j \leq k}$. The precision levels are fixed and log-normalized, defined as $\E = \left\{10^{2 - \frac{x}{5}} \right\}_{x \in [0..50]}$ (matching standard ECDF targets from~\cite{hansen2021coco}).

In Figure~\ref{fig:portofolios_total_budget}, we present two overlapping plots. In the foreground, we show the performance of the portfolios discovered by our proposed method (see Section \ref{sec:methods}) (green curve), compared to the baseline performance (red line). In the background, we visualize the composition of the portfolios for each total budget level.

We observe that certain algorithms appear consistently in the portfolio structure as the total budget increases, suggesting their general utility across a wide range of budget levels. Interestingly, only 10 algorithms contribute to portfolio formation overall, and for any given budget, at most 4 different algorithms are selected. However, some algorithms appear only within narrow budget intervals before disappearing again. While this could partially be attributed to the non-exhaustive nature of our search, it also reflects the nuanced interaction between algorithm behavior and budget allocation. This behavior provides valuable insight into the structure of the problem: it indicates that incrementally building portfolios (by extending a previous configuration as the budget grows) may fail to capture budget-specific dynamics. Certain algorithm–budget combinations are effective only within specific regimes, and a locally greedy or sequential approach may miss such opportunities. A more detailed investigation of incremental portfolio construction is presented in Section~\ref{sec:incremental}.

An additional observation is that for larger budgets, new algorithms begin to appear in the portfolio. These are often high-performing methods with restart mechanisms, and their repeated inclusion may suggest the potential benefit of tighter restart conditions or more aggressive restarts. Their presence enhances the portfolio’s diversity and adaptability as more budget becomes available.

\new{Moreover, the per-function single best solver benefits from selecting a different algorithm for each function and is therefore a stronger, function-aware baseline than the general portfolio. We investigate this setting further in Appendix~\ref{app:per-function-portfolios}, where we construct a separate portfolio for each function at a budget of \(10{,}000\) evaluations. The resulting function-specific portfolios are typically comparable to the per-function single best solvers and achieve higher performance on several functions, showing that portfolio construction remains competitive even when function-specific information is available.}

\textbf{Improvement on Individual Functions from the Set.} 
To better understand how the constructed portfolio influences individual function performance, we analyze the scenario where the total budget is fixed at $T = 10{,}000$. Specifically, we compare the performance of the portfolio against (i) the best individual algorithm for each function, and (ii) the overall mean performance across all functions. In Figure~\ref{fig:functions_best_coco}, we report the results of this comparison.

We observe that the portfolio does not outperform the best per-function algorithm. Instead, it strategically balances trade-offs across functions to achieve a stronger overall result. Certain functions (such as $F_3$ and $F_4$) show significant performance gains with the portfolio, while others like $F_{19}$, $F_{21}$, and $F_{22}$ exhibit moderate improvements. A few functions experience slight performance drops. Nonetheless, the overall average performance increases notably from \textbf{0.62} (baseline) to \textbf{0.71} with the constructed portfolio.

This effect stems from the complementary structure of the selected portfolio. When examining which algorithms contribute most to the overall performance, quantified via the normalized Shapley values shown in Figure~\ref{fig:shaply}, we observe a clear interplay between generalist and specialist strategies. A strong general-purpose algorithm, lq-CMA-ES, serves as the backbone of the portfolio, consistently delivering solid performance across the entire benchmark and closely aligning with the baseline.

Around this core, the portfolio integrates specialized algorithms that target specific performance gaps: 
\begin{itemize}
    \item BrentSTEPif (budget of 2,000 function evaluations) is highly effective on $F_3$ and $F_4$;
    \item SLSQP-11-scipy\_Hansen (budget 2,500) improves results on $F_8$, $F_9$, $F_{10}$, $F_{19}$, $F_{21}$, and $F_{22}$.
    \item lmm-CMA-ES (budget 1500) contributes strong gains on $F_{15}$, $F_{16}$, $F_{17}$, and $F_{18}$;
\end{itemize}

This composition demonstrates how the portfolio leverages algorithmic complementarity: rather than relying on one universally strong algorithm, it combines general robustness with targeted specialization to yield better overall performance.

While the experiments assume equal importance for all functions and all precision thresholds in the evaluation metric, the proposed methodology is flexible. In practice, performance metrics can be adapted to reflect user-defined priorities, such as emphasizing specific functions or prioritizing tighter precision levels. The portfolio construction strategies proposed in this work remain applicable under such modifications, enabling tailored optimization strategies for a wide range of practical scenarios.

\section{Method}\label{sec:methods}

In the previous section, we demonstrated the advantages of using budget-heterogeneous portfolios for optimizing sets of functions. However, effectively constructing such portfolios is a challenging submodular optimization task. As outlined in Section~\ref{sec:problem_formulation}, this problem is highly complex, involving a large, combinatorial search space with intricate dependencies between algorithm choices, budget allocations, and performance.

In this section, we introduce the method that was employed to construct the portfolios presented earlier. We analyze its performance across several controlled scenarios, ranging from simple to more complex settings. This analysis aims to provide a deeper understanding of both the strengths and limitations of the proposed approach.

Moreover, we have observed that the optimal portfolio structure is heavily influenced by the total available budget. Solving the problem for a smaller budget does not necessarily yield solutions that generalize to larger budgets. This complexity is further amplified by the high dimensionality of the search space, making the problem difficult not only for exact methods but also for iterative heuristic approaches (Appendix~\ref{appendix:black_box}). Algorithms designed for integer domains, such as Genetic Algorithms (Appendix~\ref{appendix:integer}), or for continuous domains, such as CMA-ES (Appendix~\ref{appendix:continuous}), often struggle to identify well-performing portfolios in this setting. Moreover, these methods are computationally expensive, as they typically require a large number of evaluations to converge. They are also prone to getting trapped in local optima, especially given the irregular and discontinuous nature of the search space.

To better understand the structure and difficulty of the search space, one possible strategy is to perform full enumeration. This approach allows us to assess the distribution and diversity of potential solutions. However, such exhaustive enumeration quickly becomes infeasible in general due to the combinatorial explosion in the number of possible portfolios, which grows exponentially with both the number of algorithms and budget options considered.

To conduct this enumeration, we begin by generating all valid sequences of budget values $\beta_i \in \vec{\beta}$ such that $\sum \beta_i \leq T$, and no additional budget from the set $\B$ can be added without exceeding the total budget $T$. Each of these sequences defines a budget composition of length $k$. For each such $\vec{\beta}$, the number of possible portfolios is given by $\size{\A}^k$, as each budget value can be independently assigned to one of the $\size{\A}$ algorithms. This exponential growth illustrates the complexity of the search space, even for relatively small values of $T$, $\B$, and $\A$.

\subsection{Greedy Local Performance Maximization}\label{sec:greedy}
\new{Let us consider an iterative process for constructing a portfolio from previously collected algorithm-performance data. At each iteration, we add a new algorithm-budget pair to the current multiset ($\MS_t$), resulting in the updated portfolio $\MS_{t+1}=\MS_t\oplus\pair{\alpha}{b}$.
}

\new{The goal is to define a selection rule for the next pair to include. A natural greedy strategy is to select the pair that maximizes the performance of the resulting portfolio:
}
\begin{equation}
    \max_{\pair{\alpha}{b} \in \A \times \B} \perf(\F, \MS_t \oplus \pair{\alpha}{b}).
\end{equation}

However, due to the monotonicity of the performance metric (i.e., adding more budget can never decrease performance  \new{, since we use anytime-algorithms where budget is not a parameter of the algorithm }), this approach will always prefer selecting the best-performing algorithm with the highest possible budget. As a result, the greedy strategy will prematurely commit to a single algorithm with budget $T$, effectively preventing any budget splitting or exploration of algorithm diversity. This limitation renders the naive greedy approach ineffective for discovering more balanced or diversified portfolio configurations.

\begin{figure}[tbp]
    \centering
    \begin{subfigure}[b]{0.48\columnwidth}
        \centering
        \includegraphics[width=\textwidth]{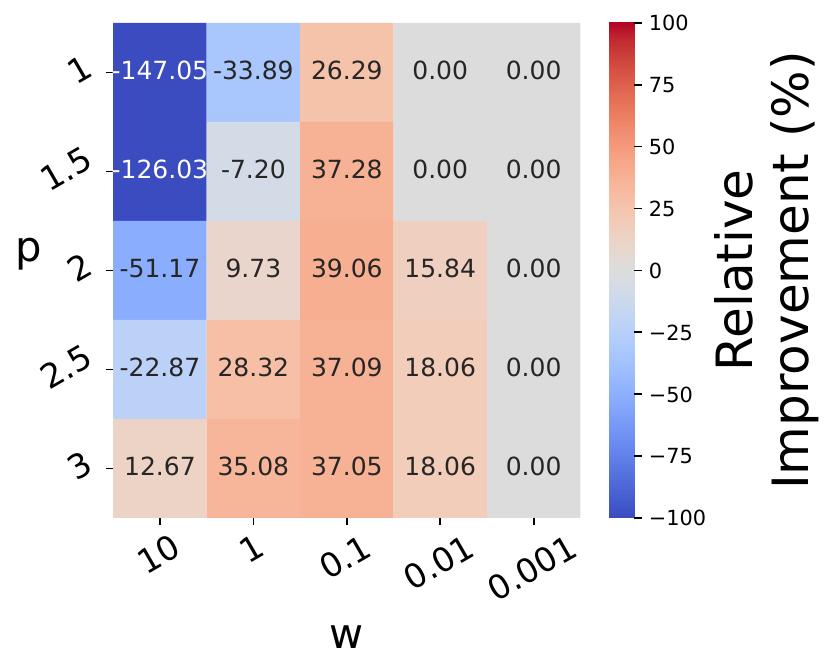}
        \caption{Relative Improvement (\%)}
        \label{fig:greedy_power_weight_improvement}
    \end{subfigure}
    \hfill
    \begin{subfigure}[b]{0.48\columnwidth}
        \centering
        \includegraphics[width=\textwidth]{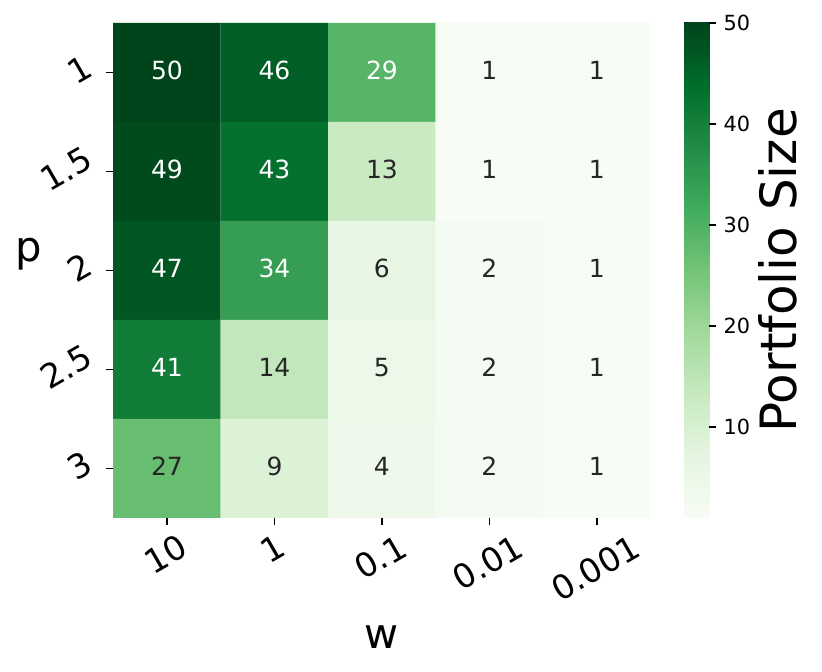}
        \caption{Portfolio Size}
        \label{fig:greedy_power_weight_size}
    \end{subfigure}
    \caption{Effect of the penalty function $\pi(b)$ on greedy portfolio construction across different weights $w$ and powers $p$.}
    \label{fig:greedy_power_weight}
\end{figure}

As such, we propose a scoring function that includes a penalty term to mildly discourage the repeated use of large budgets, without overly restricting exploration. The proposed score function is defined as:
\begin{equation}
\text{score}(t+1, \pair{\alpha}{b}) = \perf(\F, \MS_t \oplus \pair{\alpha}{b}) - \penalise(b).
\end{equation}
where $\penalise : \mathbb{N} \to \mathbb{R}$ is a penalty function applied to the budget component of the candidate pair. In our experiments, we use a simple penalty:
\begin{equation}
\penalise(b) = w  \cdot \left( \frac{b}{T} \right)^p.
\end{equation}
where both $p$ and $w$ are chosen based on a grid search displayed on Figure \ref{fig:greedy_power_weight}.

In Figure~\ref{fig:greedy_power_weight}, we use the same scenario described in Section~\ref{sec:analysis}, considering the full set of 24 functions. We observe that a \textbf{quadratic} penalization function with a weight of \textbf{0.1} yields the best performance. This function gently penalizes large budget allocations relative to the total budget $T$, while also discouraging the use of very small budgets that may be selected prematurely (when their individual contribution is minimal or redundant). This dual effect promotes a more balanced portfolio by avoiding both over-commitment to high-cost pairs and ineffective allocation of budget to low-impact configurations.

\subsection{Analysis} \label{sec:analysis}

In this section, we compare the performance of our proposed method against two baselines: the Single Best Solver (SBS) and a partial enumeration approach. These experiments also help to illustrate the inherent difficulty of the portfolio construction problem. We progressively evaluate the effectiveness of our method across increasingly complex scenarios: starting with portfolios tailored to a single function and a single algorithm, then moving to settings with two functions and three algorithms, and finally addressing the most challenging case involving all 24 BBOB functions with either one or three algorithms available.

We consider the following experimental setup:

\begin{itemize}
    \item The set of budgets is defined as $\B = \{200k \mid k \in [1..50]\}$;
    \item The total evaluation budget is $T = 10{,}000$;
    \item The precision levels are log-normalized and defined as $\E = \{10^{2-\frac{x}{10}}\}_{x\in [0..100]}$.
    \item The functions are from the BBOB suite.
\end{itemize}

We selected three optimization methods that are not necessarily the most efficient for each individual function, but rather reflect a \new{more restricted} scenario in which practitioners have access to a limited set of standard, widely available algorithms. Specifically, we include two algorithms from the scipy library \cite{2020SciPy-NMeth}: BFGS \cite{dai2002convergence} and Powell \cite{powell1964efficient}; and one from nevergrad \cite{bennet2021nevergrad}: CMA-ES \cite{auger2012tutorial}. These algorithms are representative of \new{readily available and commonly used} tools that are easy to deploy when exploring various optimization scenarios.

We evaluate the following portfolio building methods:

\begin{itemize}
    \item \textbf{Partial enumeration}: We exhaustively evaluate all valid portfolios up to a given size. For the case $\lvert\A\rvert = 1$, we enumerate portfolios of size up to $10$. For $\lvert\A\rvert = 3$, we limit enumeration to portfolios of size up to $5$, as the number of candidate solutions grows exponentially with both portfolio size and the number of algorithms.
    
    \item \textbf{Greedy method}: The greedy construction described earlier, with a scoring function that includes a mild budget penalty.
\end{itemize}
We will investigate the following scenarios:

    \begin{figure}[tbp]
        \centering
        \begin{subfigure}[b]{0.48\columnwidth}
            \centering
            \includegraphics[width=\textwidth]{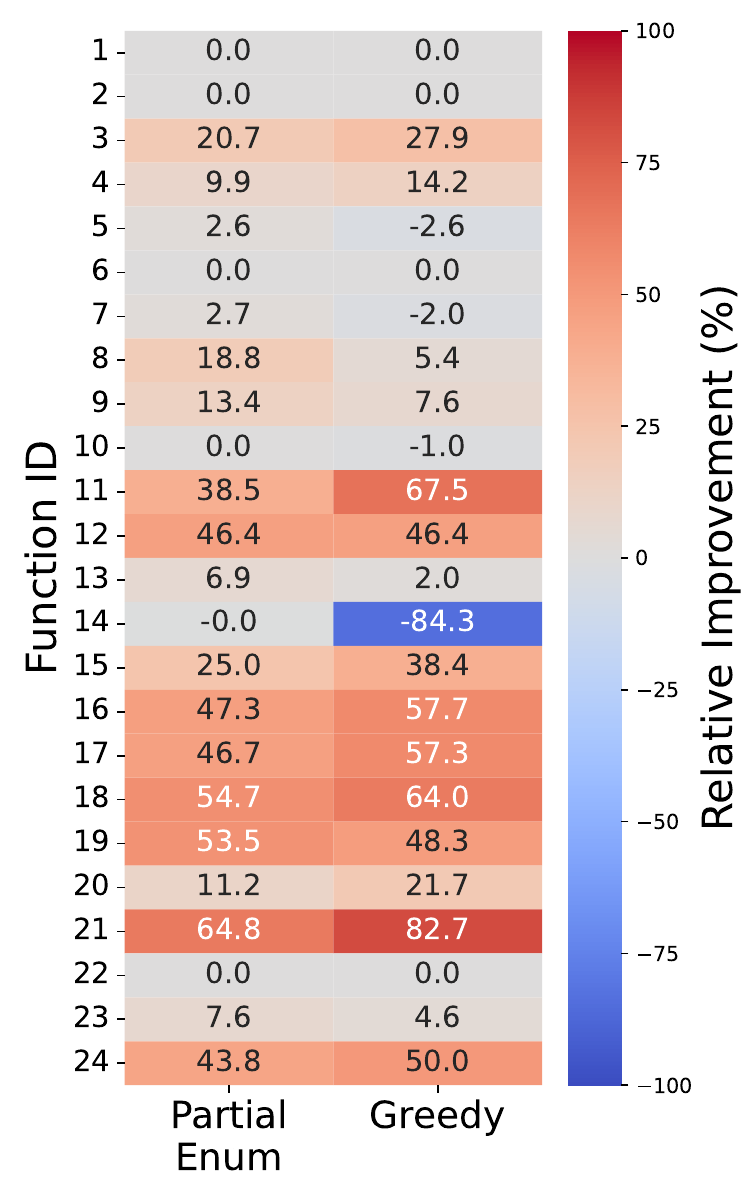}
            \caption{Relative Improvement (\%).}
            \label{fig:1a_1f_improvement}
        \end{subfigure}
        \hfill
        \begin{subfigure}[b]{0.48\columnwidth}
            \centering
            \includegraphics[width=\textwidth]{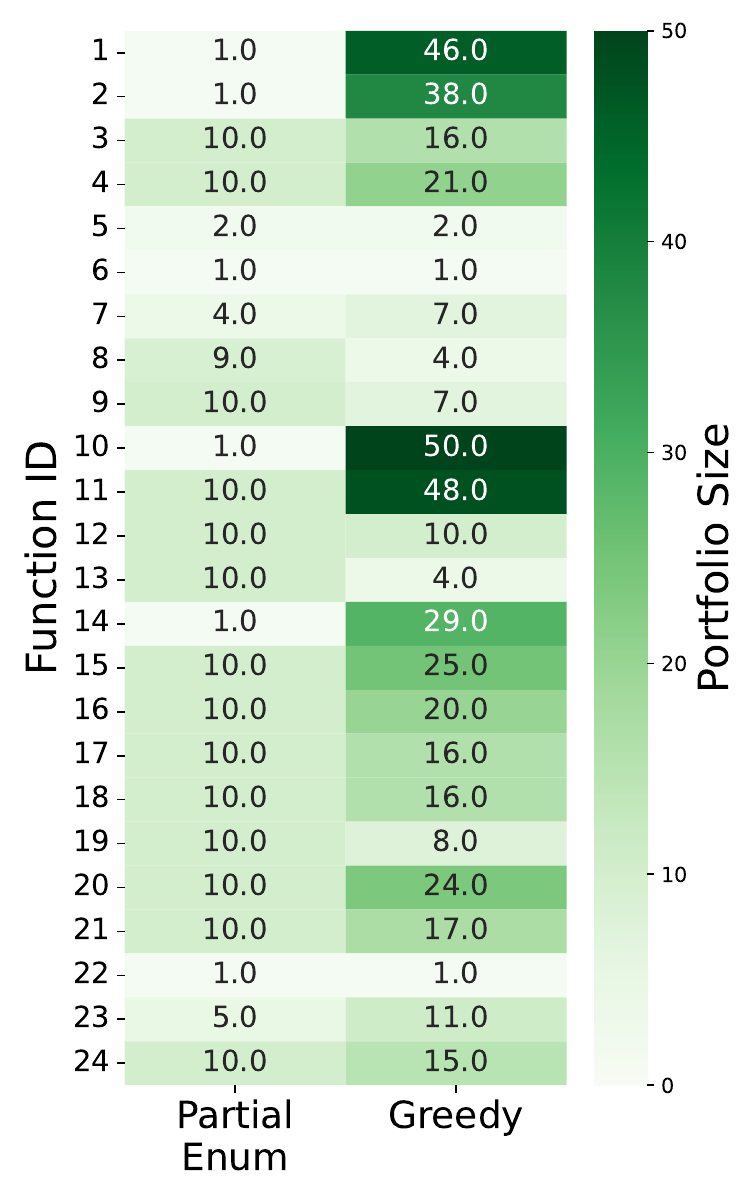}
            \caption{Average portfolio size.}
            \label{fig:1a_1f_size}
        \end{subfigure}
        \caption{Portfolio performance and size comparison for the case $\size{\A} = 1$, $\size{\F} = 1$.}
        \label{fig:1a_1f}
    \end{figure}
\textbf{1) Single Function - Single Algorithm}
    
    The first experiment evaluates the difficulty of constructing a portfolio that outperforms the baseline for the simplest setting: a single algorithm ($\A = {\text{CMA}}$) and a single function ($\size{\F} = 1$) from the BBOB suite. As shown in Figure~\ref{fig:1a_1f}, in most cases, the greedy method successfully finds portfolios that improve over the baseline. Moreover, it even outperforms the partial enumeration approach, which is limited by a maximum portfolio size. This result suggests that high-performing solutions are spread throughout the search space, making it difficult for partial enumeration (especially when constrained in size) to fully capture the optimal configurations.

    This experiment provides valuable insights into algorithm behavior. Even in the simple case of a single function and a single algorithm, splitting the budget across multiple shorter runs consistently yields better performance. This suggests that the algorithm’s internal restart mechanisms are not sufficiently aggressive (leading to inefficient budget use when runs get trapped in local minima).
    
    By distributing the budget across independent runs, the portfolio is better able to explore the search space and avoid stagnation.

\textbf{2) Two Functions - Three Algorithms}
    
\begin{figure}[tbp]
        \centering
        \begin{subfigure}[b]{0.48\columnwidth}
            \centering
            \includegraphics[width=\textwidth]{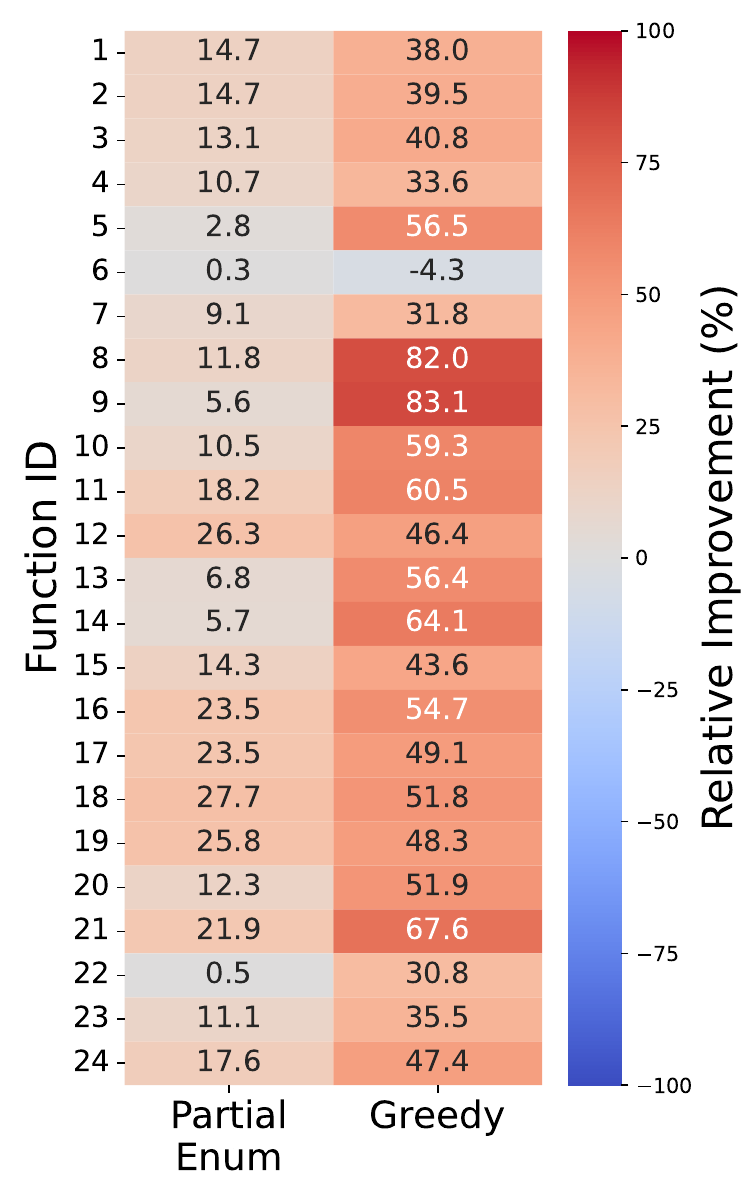}
            \caption{Relative Improvement (\%)}
            \label{fig:3a_2f_improvement}
        \end{subfigure}
        \hfill
        \begin{subfigure}[b]{0.48\columnwidth}
            \centering
            \includegraphics[width=\textwidth]{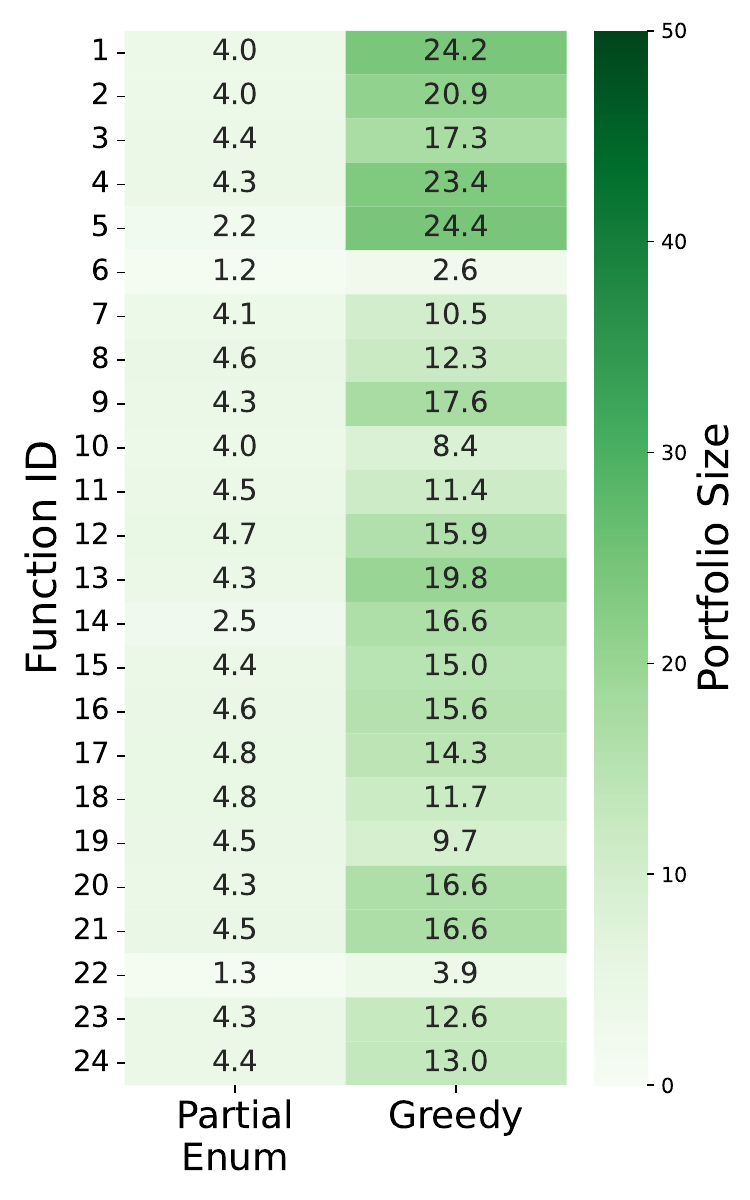}
            \caption{Average portfolio size}
            \label{fig:3a_2f_size}
        \end{subfigure}
        \caption{Portfolio performance and size comparison for the case $\size{\A} = 3$, $\size{\F} = 2$, where each function is combined with every other function}
        \label{fig:3a_2f}
    \end{figure}

    Secondly, we investigate the scenario involving pairs of functions (i.e., the simplest function sets of size two) and a set of three algorithms. In Figure~\ref{fig:3a_2f}, we show the average relative improvement obtained for all function pairs that include the function indicated on the y-axis. Compared to the single-function case, partial enumeration becomes less effective due to the exponential growth in the number of possible portfolios when more algorithms are involved.

    More importantly, we observe that the average improvement per pair is often above 30\%, highlighting the increasing value of portfolio construction in such settings. This is especially evident when comparing with previous results, suggesting that the benefits of portfolios grow significantly when more algorithms are available. This effect is largely driven by algorithm complementarity, having algorithms that specialize in different functions allows for more effective budget splitting. For example, if two algorithms each perform well on one of the two functions, it is often better to divide the budget between them than to commit to just one.

    Additionally, the observed portfolio sizes reflect not only diversification across algorithms but also internal budget splitting within each algorithm. This reinforces the insight that multiple independent runs (with varied budget allocations) can outperform single long runs, and further emphasizes the need for stronger internal restart strategies in algorithm design.
    
\textbf{3) All 24 functions - One/Three Algorithms}
    
\begin{figure}[tbp]
        \centering
        \begin{subfigure}[b]{0.48\columnwidth}
            \centering
            \includegraphics[width=\textwidth]{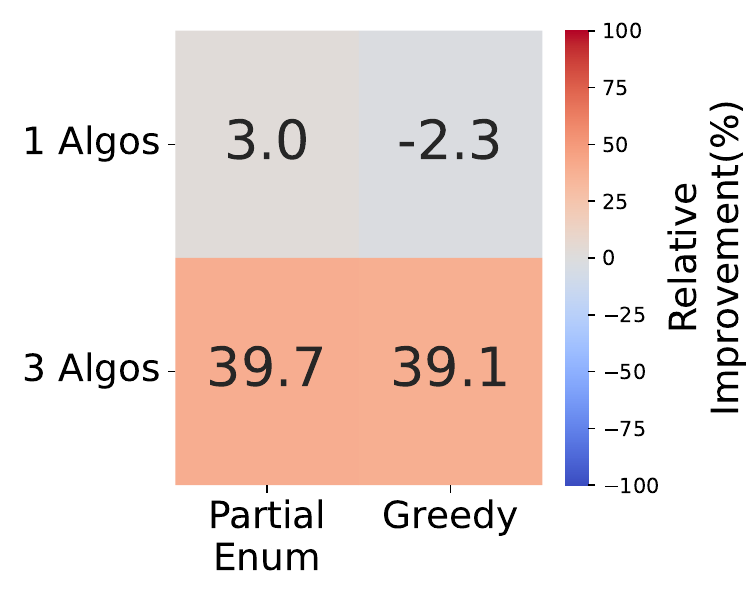}
            \caption{Relative Improvement (\%)}
            \label{fig:1_3a_24f_improvement}
        \end{subfigure}
        \hfill
        \begin{subfigure}[b]{0.48\columnwidth}
            \centering
            \includegraphics[width=\textwidth]{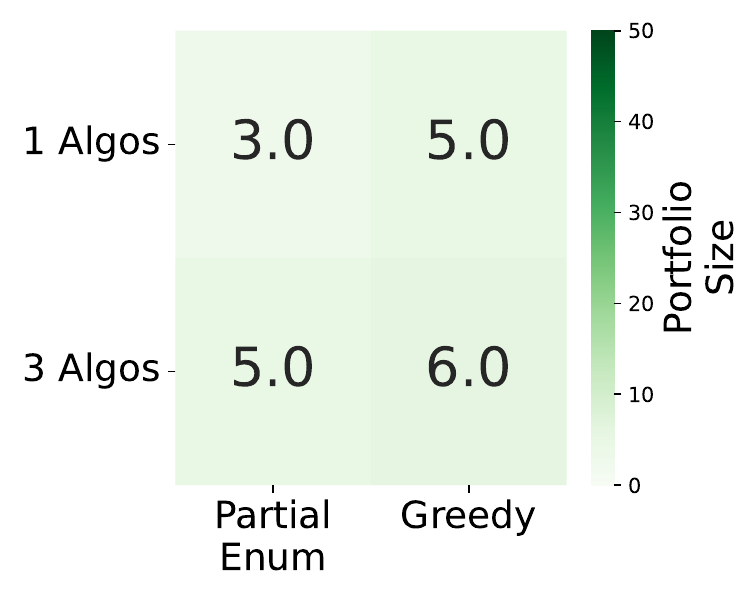}
            \caption{Average portfolio size}
            \label{fig:1_3a_24f_size}
        \end{subfigure}
        \caption{Portfolio performance and size comparison for the case of $\size{\F} = 24$.}
        \label{fig:1_3a_24f}
    \end{figure}

    Lastly, we consider the most challenging scenario: a function set consisting of all 24 BBOB functions. In Figure \ref{fig:1_3a_24f}, we compare the cases of using a single algorithm versus a portfolio of three algorithms. As expected, it becomes increasingly difficult to find effective portfolios when using only one algorithm across a diverse set of functions, each exhibiting different behaviors. In contrast, when using three algorithms, we observe improvements of up to 40\%, clearly demonstrating the power of algorithm complementarity.
   
    Moreover, as the number of functions increases, we also notice a reduction in portfolio sizes. This suggests that, in such diverse settings, allocating the budget across multiple algorithms becomes more critical. However, it also highlights that simply diversifying across algorithms is not sufficient: smaller, strategic budget splits within each algorithm (i.e., multiple independent runs with varying budgets) can significantly boost performance.
    
\new{Partial enumeration is useful for identifying recurring structures and emerging allocation patterns among promising portfolio configurations. However, it is not intended as a simple construction rule, since the set of candidate configurations must still be specified and explored in advance.}

\new{To assess whether much simpler strategies could already provide similar benefits, Appendix~\ref{app:equal-splits} considers equal budget splitting across repeated runs of a single algorithm. This baseline requires only the number of runs to be chosen, yet it is not consistently beneficial and performs worse on average than the proposed greedy construction. Extending equal splitting to multiple algorithms would introduce the additional problem of deciding which algorithm to assign to each run. These results show that simple restart-based rules are insufficient and that the algorithm selection and budget allocation performed by the proposed method are important.}

\section{Additional Experiments} \label{sec:extra}

\subsection{Utility function for Portfolio Construction}\label{sec:utility}

\begin{figure*}
    \centering
    \includegraphics[width=\linewidth]{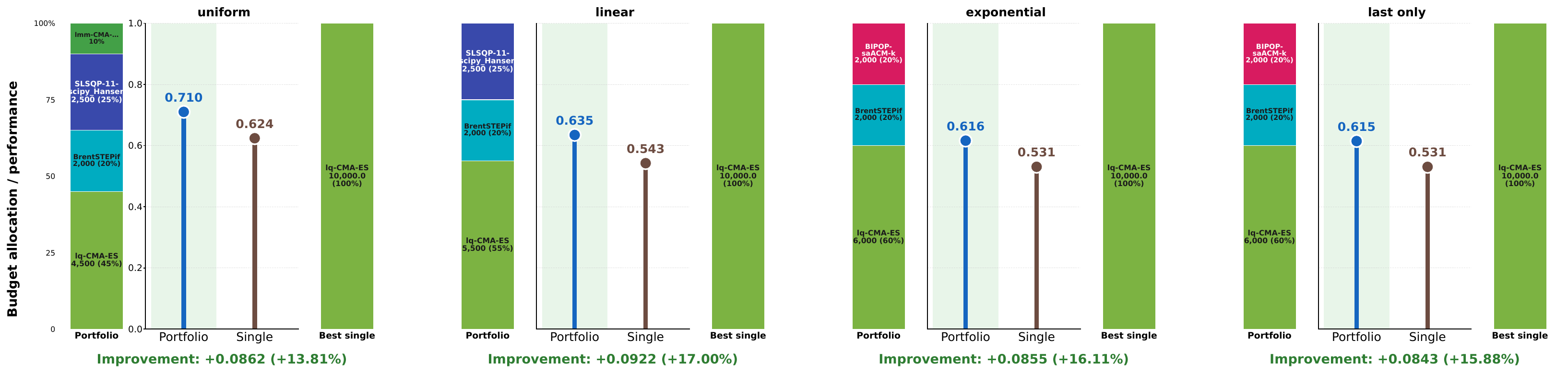}
    \caption{Portfolios constructed using four representative utility functions for a total budget of \(10{,}000\) function evaluations. For each utility, the left stacked bar shows the budget allocation of the portfolio constructed by the greedy procedure, the central panel compares its weighted performance with that of the best single algorithm, and the right bar shows the corresponding single-algorithm allocation. The portfolio outperforms the best single algorithm for all four utility profiles, including when all weight is assigned to the most demanding threshold.}
\label{fig:utility-portfolios}
\end{figure*}

\new{The EAF characterizes performance over a set of quality thresholds and evaluation budgets. Constructing a portfolio nevertheless requires these attainment probabilities to be aggregated into a scalar objective. The choice of aggregation is important, as different scoring functions can reflect different practical preferences and may consequently favor different algorithms or portfolios~\cite{graham2023formalizing}.}

\new{The performance measure in Equation~\ref{eq:performance} assigns equal importance to all precision thresholds. We generalize this objective by introducing utility weights over the thresholds and use the
resulting weighted performance measure directly within the same greedy portfolio-construction procedure. Consequently, each utility function induces a newly constructed portfolio. By comparing the resulting portfolios, we investigate whether the proposed construction method is robust to different preferences over solution quality, particularly when more demanding thresholds are assigned greater importance.}

\new{Let \(\E=\{\varepsilon_1,\ldots,\varepsilon_{\size{\E}}\}\), ordered from the easiest to the most demanding target, and let \(\U=(u_1,\ldots,u_{\size{\E}})\in[0,1]^{\size{\E}}\) be a vector of normalized utility weights satisfying}
\begin{equation}
    \sum_{i=1}^{\size{\E}} u_i = 1.
\end{equation}
\new{When harder targets are considered more valuable, the weights are chosen to be monotonically non-decreasing, such that \(u_i \leq u_{i+1}\). The resulting utility-weighted portfolio performance is defined as}
\begin{equation}
\label{eq:utility-performance}
\begin{aligned}
&\perf(\F,\MS,\U)=\\
\frac{1}{\size{\F}} \sum_{f\in\F} \color{blue}{\sum_{i=1}^{\size{\E}}} \color{blue}{u_i} &\left(1-\prod_{\pair{\alpha}{b}\in\MS}\left(1-\eaf_{f,\alpha} \bigl(b,\varepsilon_{\color{blue}{i}}\bigr)\right)\right).
\end{aligned}
\end{equation}
\new{The original performance measure is recovered by assigning equal utility to every threshold, that is, \(u_i=1/\size{\E}\) for all \(i\). Non-uniform weights allow the construction procedure to emphasize particular regions of the precision space. In the limiting case where all utility is assigned to the most demanding threshold, the objective corresponds to maximizing the probability of attaining that target.}

\begin{figure}
    \centering
    \includegraphics[width=\linewidth]{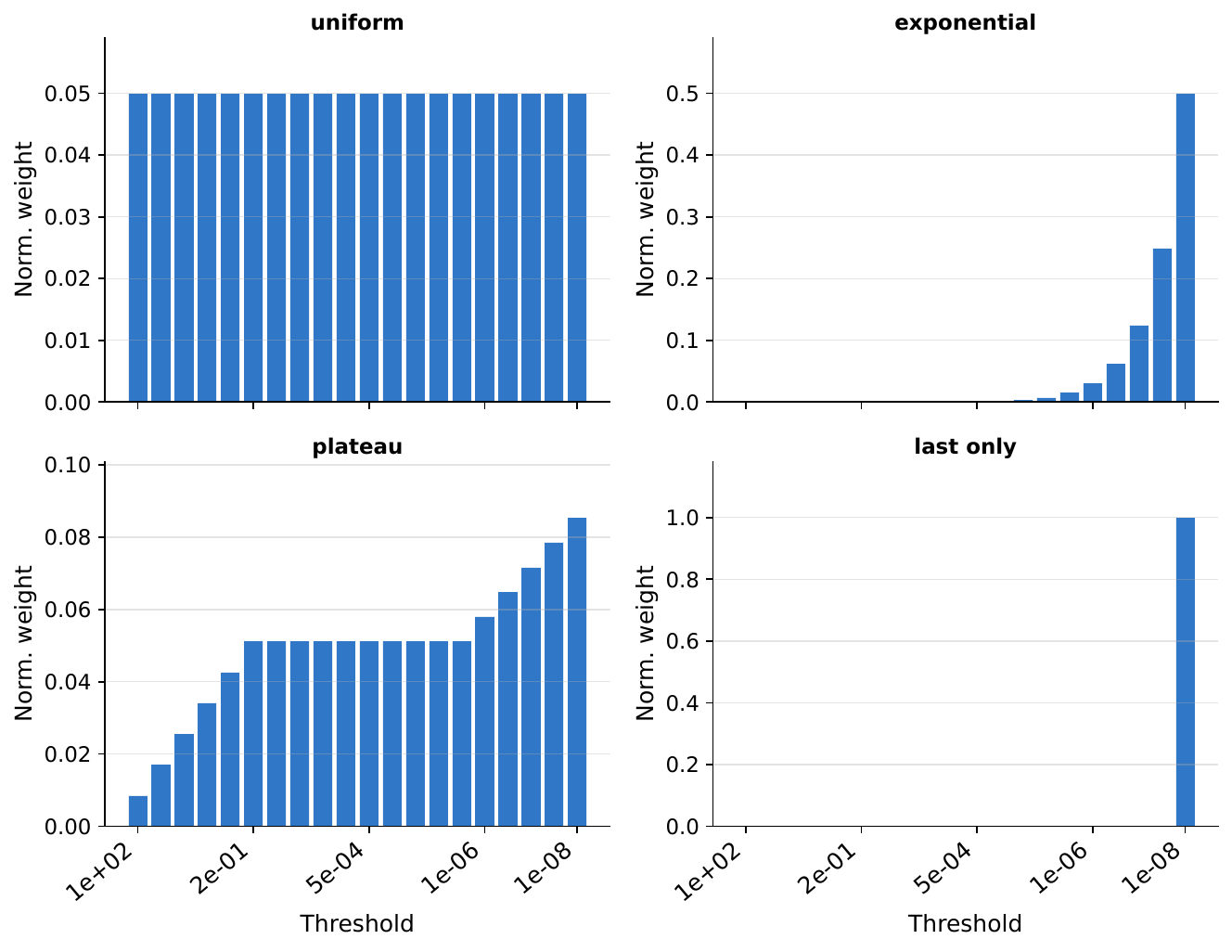}
    \caption{Representative utility-weight profiles over the precision thresholds. The thresholds are ordered from easier targets on the left to more-demanding targets on the right. Uniform weighting recovers the original performance measure, while the linear and exponential profiles assign increasing importance to harder targets. The \emph{last only} profile assigns all utility to the most demanding threshold and therefore optimizes its empirical probability of attainment. All weights are
non-negative and normalized to sum to one.}
\label{fig:utility-profiles}
\end{figure}

\new{Figure~\ref{fig:utility-profiles} illustrates four representative utility functions, ranging from uniform weighting to exclusive emphasis on the most demanding threshold. For each profile, we rerun the same greedy construction procedure using the weighted objective in Equation~\ref{eq:utility-performance}, thereby obtaining a new portfolio for each utility function.}

\new{As shown in Figure~\ref{fig:utility-portfolios}, the constructed portfolio outperforms the best single algorithm under all four utility profiles. The relative improvements range from \(13.81\%\) to \(17.00\%\), with an improvement of \(15.88\%\) when all utility is assigned to the most demanding threshold. This indicates that the method remains effective not only for average performance across thresholds, but also when the objective focuses on the probability of attaining the hardest target.}

\new{The portfolio's composition remains broadly stable, although the precise budget allocation adapts to the selected utility. Results for all twelve utility profiles are reported in Appendix~\ref{app:utility-functions}.}

\subsection{Incremental Portfolio Construction}\label{sec:incremental}

 \begin{figure*}[tbp]
        \centering
        \includegraphics[width=\linewidth]{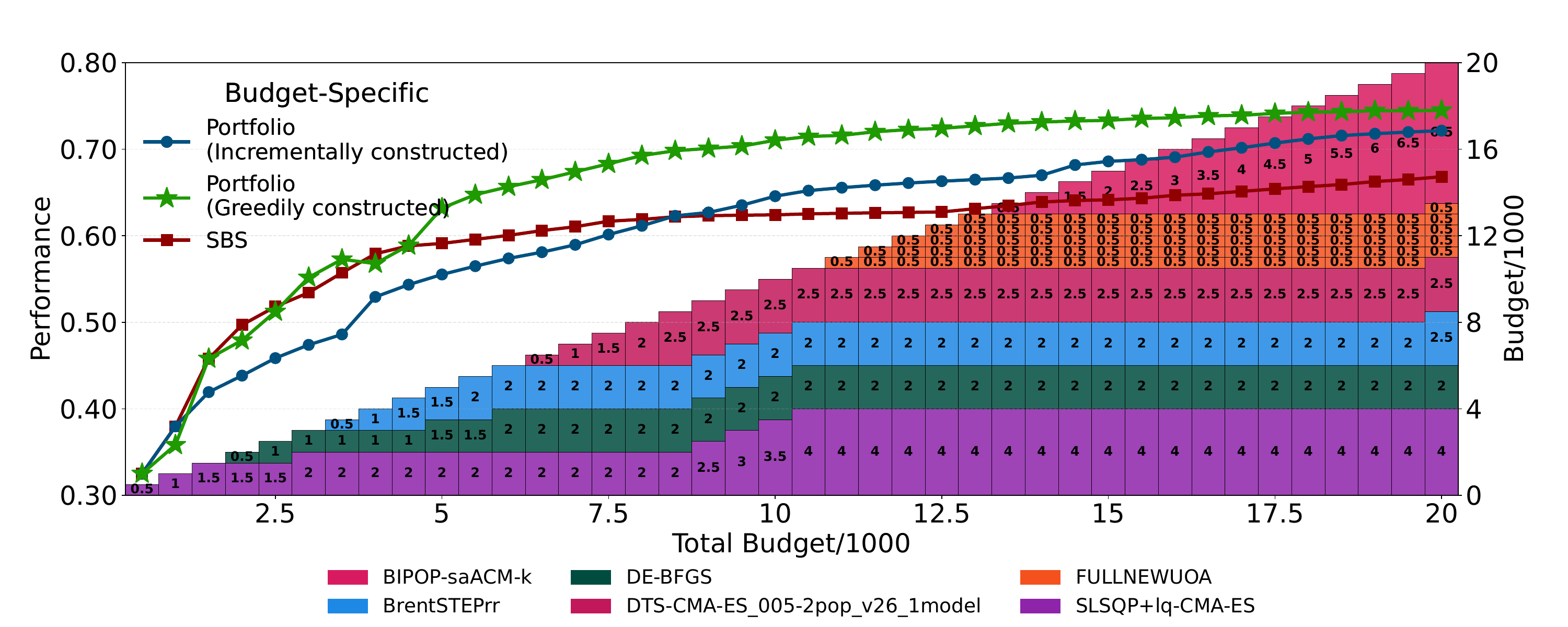}
        \caption{Performance comparison between the greedy portfolio from Fig.~\ref{fig:portofolios_total_budget}, an incremental portfolio (depicted here), and the budget-specific baseline algorithms over increasing total budgets, using a subset of COCO’s algorithm set over all 24 BBOB functions. Background elements represent the composition of the incremental portfolio. We observe that the incremental method fails to find good performing portfolio for small total budgets and it fails to pass the greedy method for higher total budgets.}
        \label{fig:incremental_portfolio}
    \end{figure*}
Up to this point, we analyzed the portfolio construction problem assuming a fixed total budget. However, another relevant scenario arises when no total budget is predefined, and the goal is to incrementally build better-performing portfolios over time.

A natural and straightforward approach in this setting is to define a fixed budget granularity and grow the portfolio step by step. Specifically, we consider a budget set $\B = \{bk \mid k \in [1..\lfloor T/b \rfloor]\}$, where $b$ is the granularity. At each stage $B_k = kb$, we take the previously constructed portfolio for $B_k$ and expand it to $B_{k+1}$ by allocating the next $b$ budget units: either to an algorithm already present in the portfolio (reinforcing its use) or to a new algorithm (increasing diversity).

In this experiment, we reuse the same setting from Section~\ref{sec:main_results}. However, instead of optimizing independently for each budget level, we grow the portfolio incrementally. In Figure~\ref{fig:incremental_portfolio}, we show the background structure of these evolving portfolios. Overlaid, we compare three performance curves: the baseline (red), the greedy per-budget portfolio (green), and the incrementally constructed portfolio (blue).

Compared to the greedy portfolios (Figure~\ref{fig:portofolios_total_budget}), we observe that incremental portfolios exhibit much lower algorithmic diversity, particularly at smaller budgets. They also consistently underperform compared to their greedy counterparts, especially in early stages. This is because, as shown previously, the portfolio composition changes significantly with budget, often in non-smooth or abrupt ways. Thus, incremental strategies may lock into suboptimal trajectories early on, making them unable to adapt to future needs.

Moreover, the choice of granularity $b$ plays a critical role. Only algorithms that yield some performance at budget $b$ can enter the portfolio early, effectively filtering out algorithms with long warm-up phases. If $b$ is too small, no algorithm may achieve measurable progress, reducing the process to random selection. Consequently, portfolios built this way are highly sensitive to granularity, and such strategies may miss high-performing solutions that only become competitive at higher budget levels.

This behavior further emphasizes the difficulty of constructing well-performing portfolios. It highlights the strong dependence on the total budget, and shows that portfolios built for smaller budgets do not generalize well to larger budgets, undermining the idea of incrementally extending good solutions.

\subsection{Influence of the Algorithm Set on Portfolio Performance}
\label{sec:random}

\begin{figure}[htbp]
    \centering
    \includegraphics[width=\linewidth]{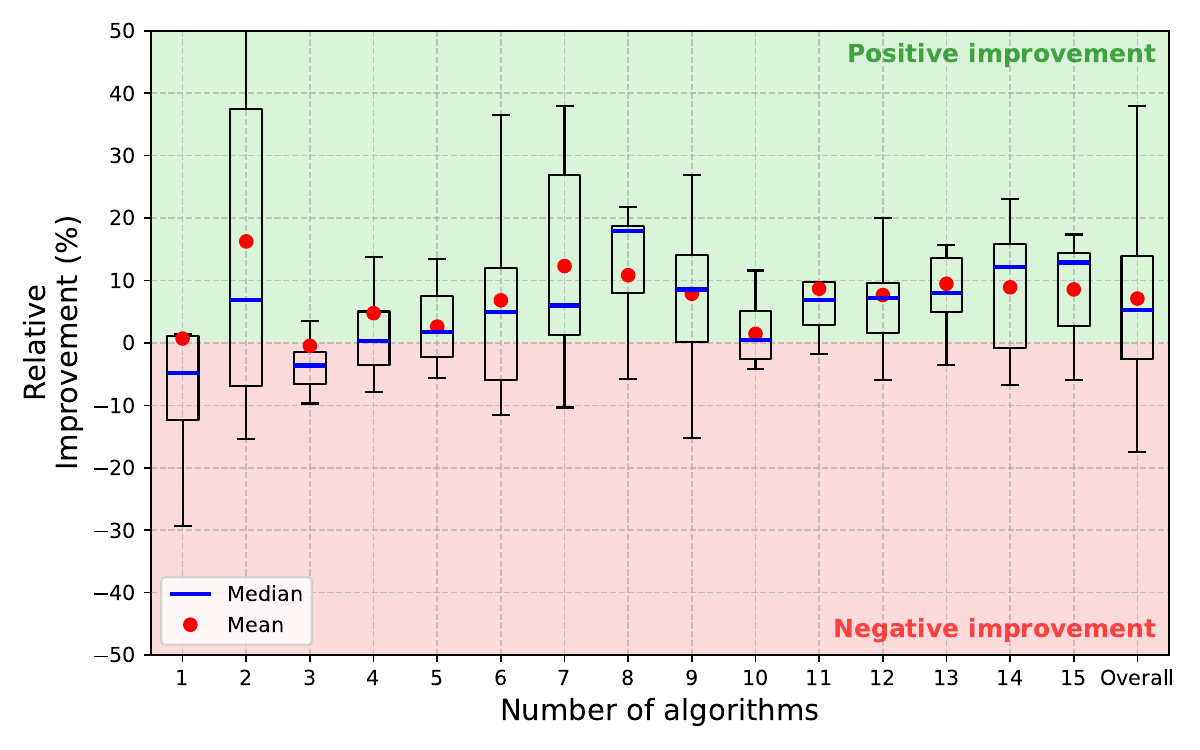}
    \caption{Improvement distribution of the greedily constructed portfolios over 10 randomly selected subsets. 
    Each box shows the median, mean (red dot), and spread of improvements relative to the baseline (SBS). Positive improvements are more frequent and stable as the number of algorithms increases.}
    \label{fig:improvement_coco_random_set}
\end{figure}
While previous sections focused on a carefully selected set of algorithms, we now aim to evaluate our method in a more realistic and large-scale context. To this end, we consider the same COCO dataset and the scenario introduced in Section~\ref{sec:main_results}. Specifically, we investigate how the composition of the algorithm set $\mathcal{A}$ influences the portfolio performance; examining whether, in a fully randomized setting, portfolios still achieve a net positive improvement over the baseline.

In this experiment, we randomly sample subsets of algorithms of varying sizes ($|\mathcal{A}| \in [1, 15]$), drawing multiple sets for each size to ensure statistical robustness. For each sampled set, we apply our greedy portfolio construction method and compare its performance against a simple baseline portfolio.

Figure~\ref{fig:improvement_coco_random_set} shows the distribution of improvements across these trials. We observe that improvements are easier to obtain as the number of available algorithms increases. On average, our method achieves a relative improvement of 5–10\% over the baseline, regardless of algorithm set size.

Although these improvements are moderate, the low computational cost of the greedy approach makes it particularly attractive in practical settings. This method can serve not only as an efficient standalone strategy, but also as a strong initialization point for more expensive local search heuristics, such as a $(1+1)$ Genetic Algorithm seeded with the greedy portfolio.

\section{Conclusions and Future Work}

This work demonstrates the effectiveness of sequential budget-heterogeneous algorithm portfolios as an alternative to traditional single-algorithm selection strategies. Across a range of scenarios (varying in complexity, number of functions, and algorithm availability), we observed consistent gains in performance when using portfolios, particularly due to the complementary strengths of generalist and specialist algorithms.

However, constructing effective portfolios remains a challenging problem. The underlying search space is combinatorially large and highly sensitive to both the total budget and the structure of the benchmark set. While our greedy method provides a scalable and effective baseline, default black-box optimization techniques often struggle to match its performance, particularly due to local minima and search space fragmentation.

\new{A particularly interesting observation is the consistent benefit of distributing the available budget across multiple shorter runs of the same algorithm, even in the single-function setting. Under the evaluated conditions, this strategy appears to provide greater robustness than a single long run. While premature convergence or limitations in existing restart mechanisms may help explain this behavior, the present experiments do not directly test these hypotheses. Further work is therefore needed to determine the underlying causes and to assess whether more adaptive restart strategies can systematically improve performance.}

\new{Another interesting natural direction is to integrate portfolio methods into algorithm selection frameworks. Unlike standard selection mechanisms that commit to a single algorithm, portfolios can inherently account for uncertainty in function or feature space, potentially leading to more robust behavior under uncertainty. Recent work on similarity-based portfolio construction, building on top of the methods discussed in this paper, has shown promising results in fixed-budget black-box optimization, demonstrating that portfolios can outperform traditional single-solver selection baselines \cite{dinu2026similarity}.}

\new{Although our portfolios assume sequential and independent evaluation, they open up several directions for more advanced coordination strategies. One natural extension is algorithm scheduling, where the order and timing of portfolio components could be adapted rather than fixed in advance. Beyond scheduling, a further avenue is to consider knowledge transfer between portfolio components, for example by using the outcome of one algorithm run as a warm start or initialization for the next. Such mechanisms could lead to stronger sequential portfolios, provided care is taken to avoid overfitting or narrowing the search space too early. Designing ``soft chaining'' strategies, where limited information is shared without full state transfer, may offer a way to combine the robustness of independent portfolios with the potential benefits of coordination.}

Finally, these methods could be embedded into higher-level automated systems, such as the ngopt wizard~\cite{meunier2021black} in Nevergrad, or standard AutoML frameworks, offering practical benefits in real-world tuning and adaptation pipelines.
\bibliographystyle{IEEEtran}
\bibliography{bibliography}

\appendices
\section{\new{Function-Specific Portfolios}}\label{app:per-function-portfolios}

\new{To investigate the potential benefit of specialization, we construct one portfolio for each BBOB function at a fixed total budget of \(10{,}000\) function evaluations. The candidate algorithms, precision thresholds, and greedy construction procedure remain unchanged. The only difference is that the objective is evaluated separately for each function.}

\begin{figure*}[t]
    \centering
    \includegraphics[width=\textwidth]{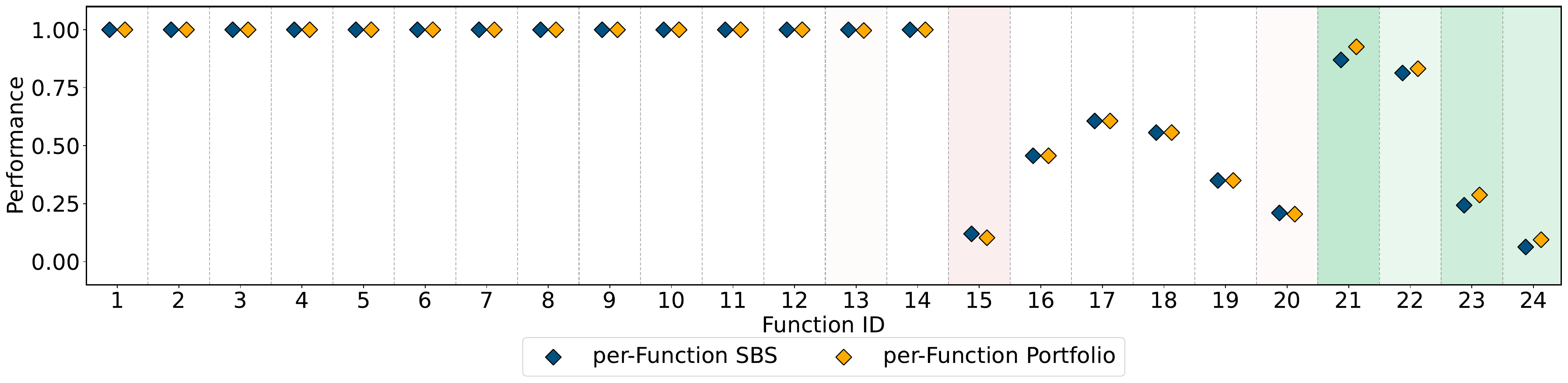}
    \caption{Performance of the per-function single best solver and the function-specific portfolio on each BBOB function at a fixed budget of \(10{,}000\) function evaluations. A separate portfolio is constructed for every function using the same greedy procedure. Green shaded regions indicate functions for which the portfolio outperforms the per-function single best solver, while red shaded regions indicate functions for which the portfolio performs worse.}
    \label{fig:per-function-portfolios}
\end{figure*}

\new{Figure~\ref{fig:per-function-portfolios} compares the resulting function-specific portfolios with the single best solver selected independently for each function. On Functions \(1\)-\(14\), both approaches achieve performance close to one, leaving little room for improvement. On the remaining functions, the results are more varied. The function-specific portfolio improves over the per-function single best solver on the functions highlighted in green, while it performs worse on the functions highlighted in red. These differences indicate that portfolio construction can exploit complementary behavior on some functions, but does not guarantee an improvement for every individual function.}

\section{\new{Additional Utility Functions}}\label{app:utility-functions}

\begin{figure}
    \centering
    \includegraphics[width=\linewidth]{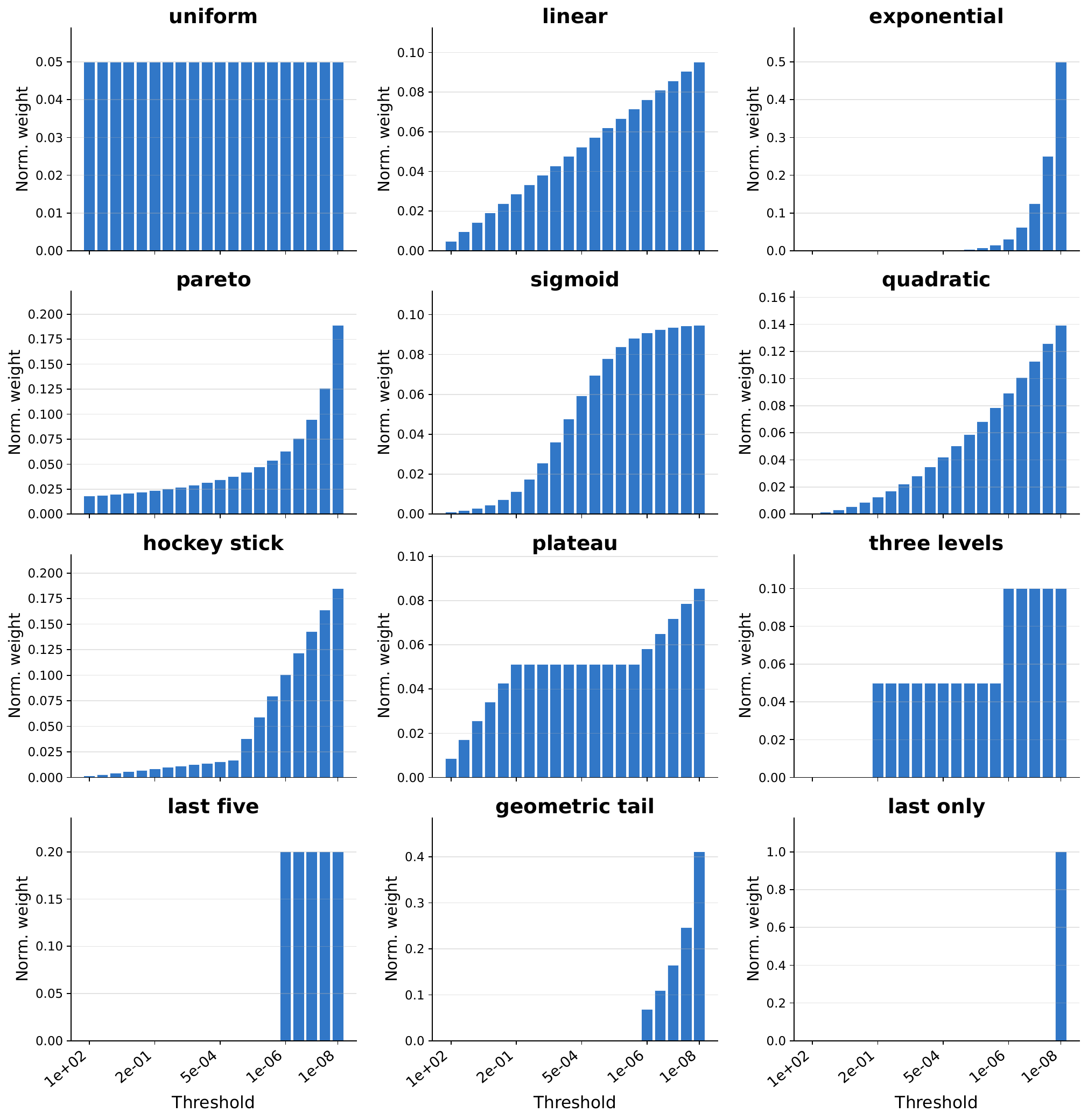}
    \caption{The twelve normalized utility-weight profiles considered in the sensitivity analysis. Thresholds are ordered from easier targets on the left to more demanding targets on the right. The profiles range from uniform weighting to assigning all utility to the most demanding threshold. Each profile is normalized such that its weights sum to one.}
    \label{fig:all-utility-profiles}
\end{figure}

\begin{figure*}
    \centering
    \includegraphics[width=\linewidth]{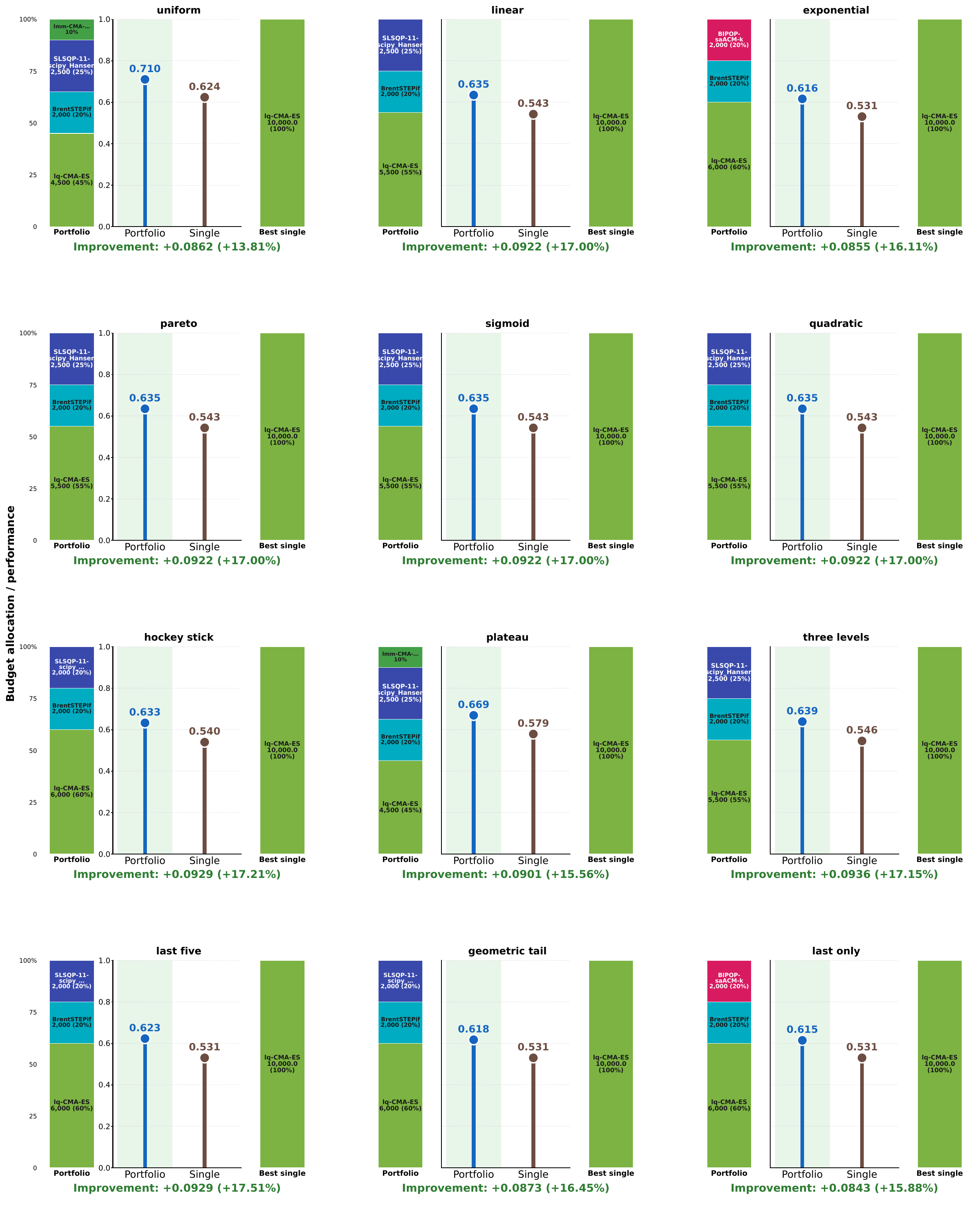}
    \caption{Portfolios constructed under the twelve utility-weight profiles using a total budget of \(10{,}000\) function evaluations. For each utility, the left stacked bar shows the portfolio budget allocation, the central panel compares its weighted performance with that of the best single algorithm, and the right bar shows the corresponding single-algorithm allocation. A new portfolio is constructed for each utility using the same greedy procedure with the weighted objective. The portfolio outperforms the   best single algorithm for every utility profile.}
\label{fig:all-utility-portfolios}
\end{figure*}

\new{To investigate the sensitivity of the portfolio-construction method to the aggregation of attainment probabilities, we consider twelve normalized utility profiles over the precision thresholds. The profiles cover both smooth weighting schemes and highly selective objectives. For every profile, the utility weights satisfy:}
\begin{align}
    u_i \geq 0, \\
    \sum_{i=1}^{\size{\E}} u_i = 1
\end{align}
\new{The thresholds are ordered from easiest to most demanding, such that larger indices correspond to harder targets.}
\new{
The considered utility profiles are:
\begin{enumerate}
    \item \textbf{Uniform:} assigns equal utility to all precision thresholds and recovers the original performance measure.
    \item \textbf{Linear:} increases the utility linearly with target difficulty.
    \item \textbf{Exponential:} concentrates utility increasingly strongly on the most demanding thresholds.
    \item \textbf{Pareto:} assigns gradually increasing utility, with a stronger emphasis on the hardest targets.
    \item \textbf{Sigmoid:} provides a smooth transition from low utility for easier targets to high utility for more demanding targets.
    \item \textbf{Quadratic:} increases the utility quadratically with target difficulty.
    \item \textbf{Hockey stick:} assigns relatively little utility to most thresholds and increases sharply for the most demanding ones.
    \item \textbf{Plateau:} increases initially, remains approximately constant over an intermediate range, and increases again for the hardest targets.
    \item \textbf{Three levels:} assigns zero utility to easier targets, moderate utility to intermediate targets, and higher utility to the most demanding targets.
    \item \textbf{Last five:} distributes utility uniformly over only the five most demanding thresholds.
    \item \textbf{Geometric tail:} concentrates utility on the most demanding thresholds according to a geometric progression.
    \item \textbf{Last only:} assigns all utility to the single most demanding threshold.
\end{enumerate}
}
\new{
Figure~\ref{fig:all-utility-profiles} shows the complete set of weighting
profiles. For every utility, the greedy construction procedure is run
again using the corresponding weighted objective. The resulting
portfolios and their comparison with the best single algorithm are shown
in Figure~\ref{fig:all-utility-portfolios}.
}

\section{Treating the Performance Metric as a Black-Box Optimization Task}
\label{appendix:black_box}
In this context, we treat the portfolio optimization task as a black-box optimization problem, where the performance metric $\perf(\F, \MS)$ acts as an objective function that is expensive to evaluate and lacks analytical gradient information. This setting naturally arises when evaluating the performance of a portfolio requires running multiple stochastic algorithms on a diverse set of benchmark functions.

As such, we propose two alternative representations for solutions to the portfolio problem, along with several algorithmic strategies designed to operate effectively on these representations. These representations allow us to leverage both discrete and continuous optimization methods to explore the combinatorial space of portfolio configurations under a budget constraint.

\subsection{Integer Search Space}\label{appendix:integer}

To work with a compact and optimization-friendly representation, we introduce the \emph{multiplicity function} specific to the multiset $\MS$, denoted as:
\begin{equation}
    \multi : \A \times \B \rightarrow \mathbb{N}
\end{equation}

This function specifies how many times each pair $\pair{\alpha}{b}$ from the Cartesian product $\A \times \B$ appears in the multiset $\MS$.

We can equivalently represent this multiset as an \emph{integer vector} $\vec{x} \in \mathbb{N}^{\left(|\A| \cdot |\B|\right)}$, where each component corresponds to the multiplicity of a specific algorithm-budget pair:
\begin{equation}
    \vec{x} = \{x_{\pair{\alpha}{b}}\}_{\pair{\alpha}{b}\in\A \times \B}, \text{ with  } x_{\pair{\alpha}{b}} = \multi(\pair{\alpha}{b})
\end{equation}
Using the integer vector representation introduced earlier, we can express the budget constraint as:  
\begin{equation}
    \sum_{\pair{\alpha}{b}\in\A\times\B} x_{\pair{\alpha}{b}} b \leq T
\end{equation}
and rewrite the performance function $\perf(\F, \vec{x})$ as:

\begin{equation}
    \frac{1}{|\F| \cdot |\E|} \sum_{f \in \F}
\sum_{\varepsilon \in \E} \left( 1 - \prod_{\pair{\alpha}{b} \in \A \times \B}
\left( 1 - \eaf_{f,\alpha}(b, \varepsilon) \right)^{x_{\pair{\alpha}{b}}} \right)
\end{equation}

This representation naturally lends itself to the application of population-based metaheuristics such as Genetic Algorithms (GAs). In our implementation, we enforce that all candidate solutions remain valid, i.e., they must satisfy the total budget constraint. After crossover and mutation, any offspring violating the constraint are \emph{repaired} by decrementing components of $\vec{x}$ until feasibility is restored.

For the \emph{mutation operator}, we apply a simple mechanism: with a fixed mutation probability, each gene $x_{\pair{\alpha}{b}}$ can be either incremented or decremented by 1, subject to non-negativity.

For the \emph{crossover operator}, any integer-based crossover can be applied. This includes standard operators such as $n$-point crossover, uniform crossover, or element-wise operations like mean crossover (rounding as necessary to preserve integrality).
\subsection{Continuous Search Space}\label{appendix:continuous}
In the continuous search space setting, we first consider how to optimally distribute the total budget among a set of algorithms. One simple and intuitive approach is to assign a fixed proportion of the total budget to each algorithm. For example, allocating $40\%$ to the first algorithm, $50\%$ to the second, and $10\%$ to the third corresponds to the portfolio:
\begin{equation}
    \MS^{*}\{
    (\alpha_1,\lfloor 0.40 \cdot T \rfloor)\ 
    (\alpha_2, \lfloor 0.50 \cdot T \rfloor)\ 
    (\alpha_3,\lfloor 0.10 \cdot T \rfloor))
    \}
\end{equation}

This representation allows us to explore the entire space of budget allocations among algorithms, enabling exhaustive evaluation of all possible proportional configurations. However, while this proportional representation provides valuable insight into how budget should be divided between algorithms, it overlooks a large portion of the search space. Specifically, scenarios where the budget assigned to an individual algorithm is also split internally (i.e., used at different budget levels).
To capture this broader class of solutions, we introduce a constraint on the maximum number of splits allowed per algorithm, denoted $k$. Each algorithm can thus be assigned up to $k$ different budget levels. We define a continuous solution vector of size $k * |\A|$:
\begin{align*}
    \vec{x} = \{x_{\alpha,i}\}_{\alpha\in\A, 0\leq i < k},\\
    \text{s.t. } \sum_{\alpha\in\A}\sum_{i=0}^{k} x_{\alpha,i} = 1
\end{align*}

Each element $x_{\alpha,i}$ represents a fraction of the total budget to be assigned to the $i$-th sub-budget level for algorithm $\alpha$. This representation allows fine-grained encoding of both inter-algorithm and intra-algorithm budget allocation, and can be decoded into valid portfolios accordingly. 

In order to enable the use of continuous optimizers for solving the portfolio problem, we require a solution representation that satisfies the budget constraint. While this constraint may appear tight, it is possible to transform any continuous solution into a valid one through appropriate normalization.

We consider a raw solution vector $\vec{y} \in [0,1]^{k \cdot |\A|}$, indexed as $y_{\alpha,i}$, where $\alpha \in \A$ and $i = 1, \dots, k$. This vector is then mapped to a normalized vector $\vec{x} \in \Delta^{k \cdot |\A| - 1}$, indexed similarly as $x_{\alpha,i}$, using one of the following strategies:

\begin{enumerate}
    \item \textbf{Simple normalization}:
    \begin{equation}
        x_{\alpha,i} = \frac{y_{\alpha,i}}{\sum_{\alpha \in \A} \sum_{i=1}^{k} y_{\alpha,i}}
    \end{equation}
    This ensures that $\vec{x}$ lies on the probability simplex, i.e., all components are non-negative and sum to one.

    \item \textbf{Dirichlet-based stick-breaking transformation} \cite{paisley2010simple}:

    Define $D = k \cdot |\A|$ and let $\vec{y} \in [0,1]^{D-1}$. We compute an intermediate vector $\vec{t} \in \mathbb{R}^{D-1}$ as follows:
    \begin{align*}
        t_1 &= \mathrm{Beta}^{-1}(y_1;\ 1,\ D) \\
        t_i &= t_{i-1} + (1 - t_{i-1}) \cdot \mathrm{Beta}^{-1}(y_i;\ 1,\ D - i + 1)
    \end{align*}
    The final normalized components are:
    \begin{equation}
        x_1 = t_1,\quad x_i = t_i - t_{i-1},\quad x_D = 1 - t_{D-1}
    \end{equation}
    These values can be reshaped to obtain $x_{\alpha,i}$ as needed. This procedure guarantees that $\vec{x} \in \Delta^{D-1}$.

    \item \textbf{Euclidean projection onto the simplex} \cite{wang2013projection}:

    The algorithm proceeds as follows:
    \begin{enumerate}
        \item Sort the components of $\vec{y}$ in descending order to obtain $\vec{u}$.
        \item Compute the cumulative sum $\text{cssv} = \sum_{j=1}^i u_j - 1$.
        \item Find the largest index $\rho$ such that $u_\rho > \frac{1}{\rho} \left( \sum_{j=1}^\rho u_j - 1 \right)$.
        \item Compute the threshold $\theta = \frac{1}{\rho} \left( \sum_{j=1}^\rho u_j - 1 \right)$.
        \item Set $x_i = \max(y_i - \theta,\ 0)$.
    \end{enumerate}
    This yields a vector $\vec{x}$ that lies on the simplex and is closest to $\vec{y}$ in the Euclidean sense.
\end{enumerate}
Considering the same scenario as in Section~\ref{sec:analysis}, we now investigate how the parameter $k$ and the choice of normalization method affect the performance of the optimizer. As the optimization method, we use the Powell algorithm from SciPy.

\begin{figure}[tbp]
    \centering
    \begin{subfigure}[b]{0.48\columnwidth}
        \centering
        \includegraphics[width=\textwidth]{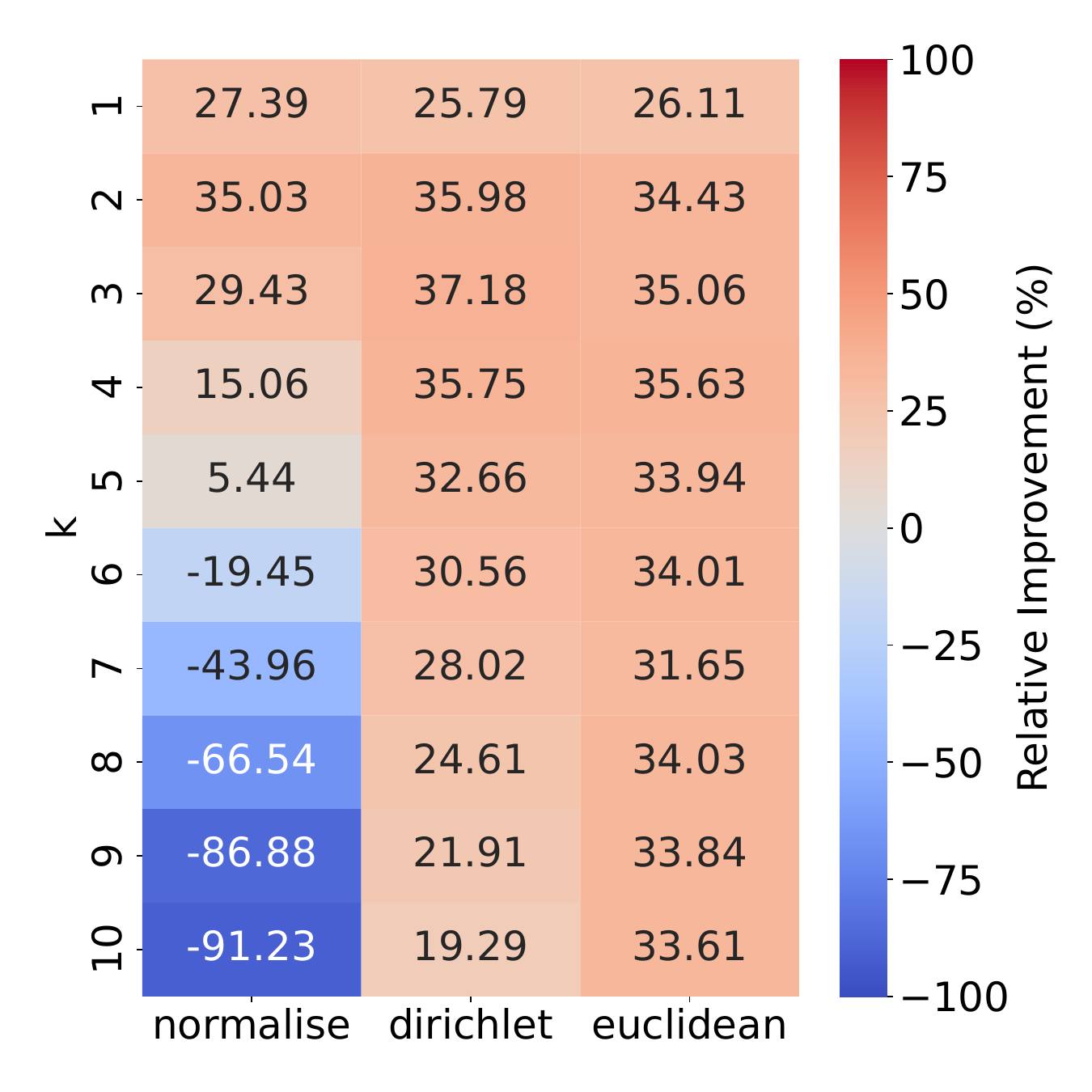}
        \caption{Relative Improvement (\%)}
        \label{fig:k_influence_improvment}
    \end{subfigure}
    \hfill
    \begin{subfigure}[b]{0.48\columnwidth}
        \centering
        \includegraphics[width=\textwidth]{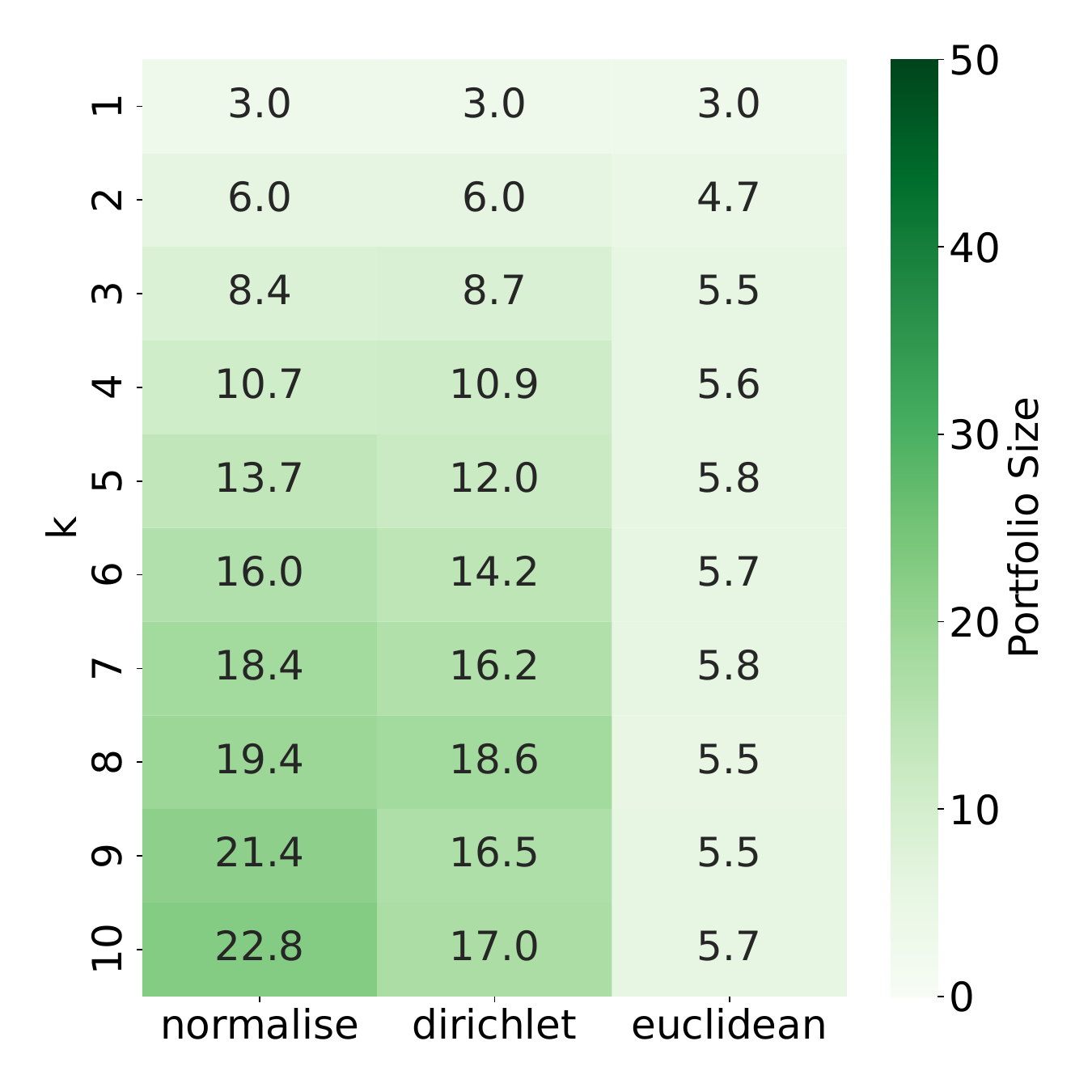}
        \caption{Average Portfolio Size}
        \label{fig:k_influence_size}
    \end{subfigure}
    \caption{Comparison of normalization methods (Simple, Dirichlet, Euclidean) across different values of $k$. }
    \label{fig:k_influence}
\end{figure}

Figure~\ref{fig:k_influence_improvment}  shows the relative improvement achieved for each combination of $k$ and normalization method, while the Figure \ref{fig:k_influence_size} reports the corresponding portfolio sizes. We observe that simple normalization becomes unstable as $k$ increases, leading to volatile performance and unpredictable portfolio sizes. In contrast, Euclidean projection offers consistent performance across different values of $k$, producing stable and compact portfolios.

However, the method that strikes the best balance between stability and performance is the Dirichlet-based projection. It consistently yields high performance and remains robust across settings, with the best result obtained at $k=3$. Therefore, for all subsequent continuous optimization experiments, we adopt Dirichlet projection with $k=3$ as the default strategy.
 
\section{Full Analysis}\label{app:full_analysis}
In this section, we compare the performance of the various methods introduced in the previous sections. We consider the experimental setup from section \ref{sec:analysis} and we evaluate the following methods:
\begin{enumerate}
    \item \textbf{Partial enumeration}: We exhaustively evaluate all valid portfolios up to a given size. For the case $\lvert\A\rvert = 1$, we enumerate portfolios of size up to $10$. For $\lvert\A\rvert = 3$, we limit enumeration to portfolios of size up to $5$, as the number of candidate solutions grows exponentially with both portfolio size and the number of algorithms.
    
    \item \textbf{Greedy Portfolio} (section \ref{sec:greedy}) with a scoring function that includes a mild budget penalty.

    \item \textbf{Incremental Portfolio} (section \ref{sec:incremental})    
    \item \textbf{Genetic algorithms}: We consider two evolutionary setups: $(1+1)$ for local search behavior, and $(10+20)$ for a population-based strategy.
    
    \item \textbf{Continuous optimization methods}: We use Dirichlet projection with $k=3$ as discussed in Section~\ref{appendix:continuous}. Two optimizers are tested: Powell (local search) and CMA-ES (population-based).
\end{enumerate}

Each heuristic method is allowed a total computational budget of $1000 \cdot \size{\A}$ fitness evaluations.

\subsection{Single Function - Single Algorithm}
We begin by analyzing the simplest possible case: a portfolio constructed using only a single algorithm ($\size{\A} = 1$) and optimized for a single function ($\size{\F} = 1$). In this setting, we investigate whether splitting the total evaluation budget across multiple shorter runs can outperform allocating the entire budget to a single long run.
\begin{figure}[tbp]
    \centering
    \begin{subfigure}[b]{0.48\columnwidth}
        \centering
        \includegraphics[width=\textwidth]{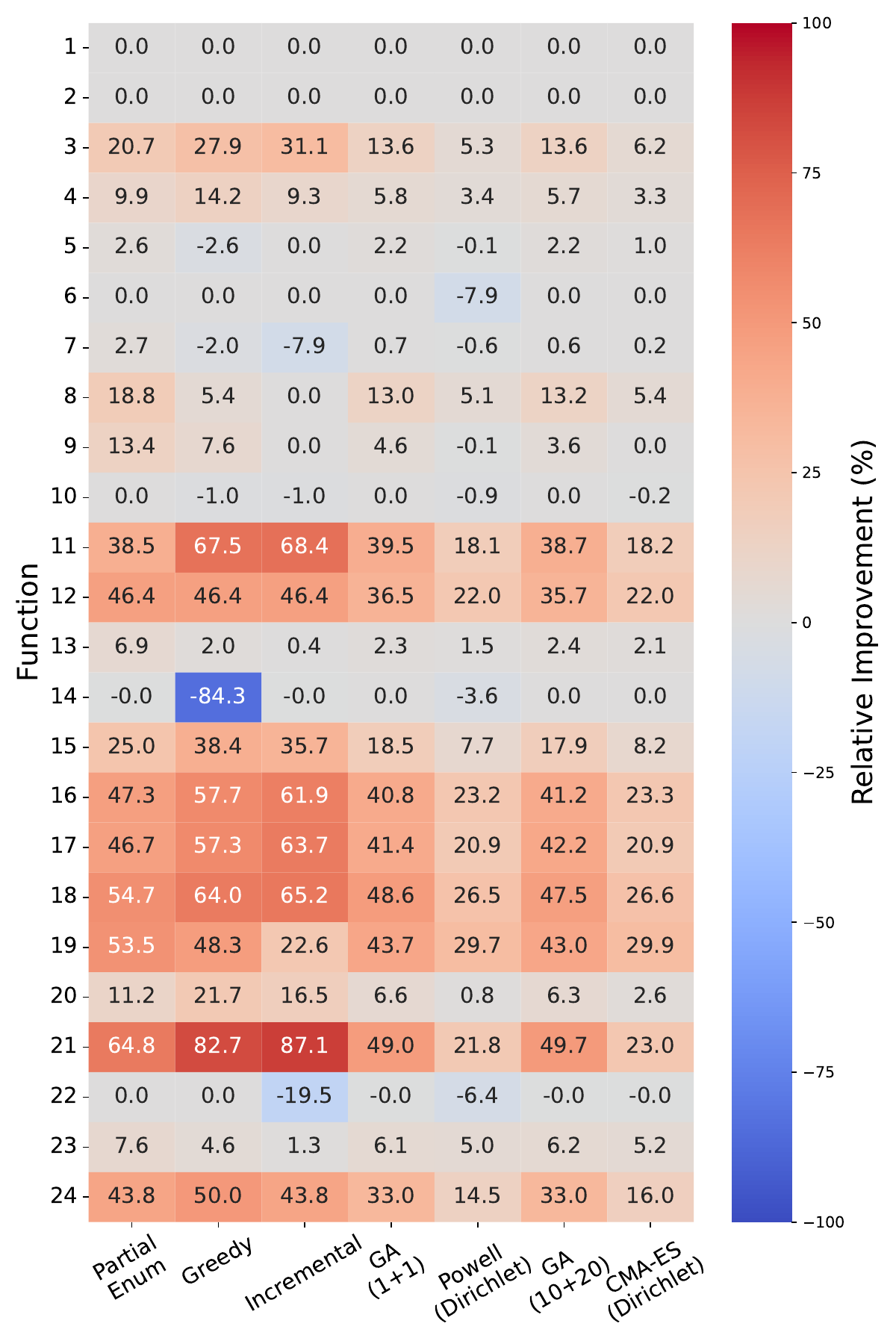}
        \caption{Relative Improvement (\%).}
        \label{fig:1a_1f_appendix_improvement}
    \end{subfigure}
    \hfill
    \begin{subfigure}[b]{0.48\columnwidth}
        \centering
        \includegraphics[width=\textwidth]{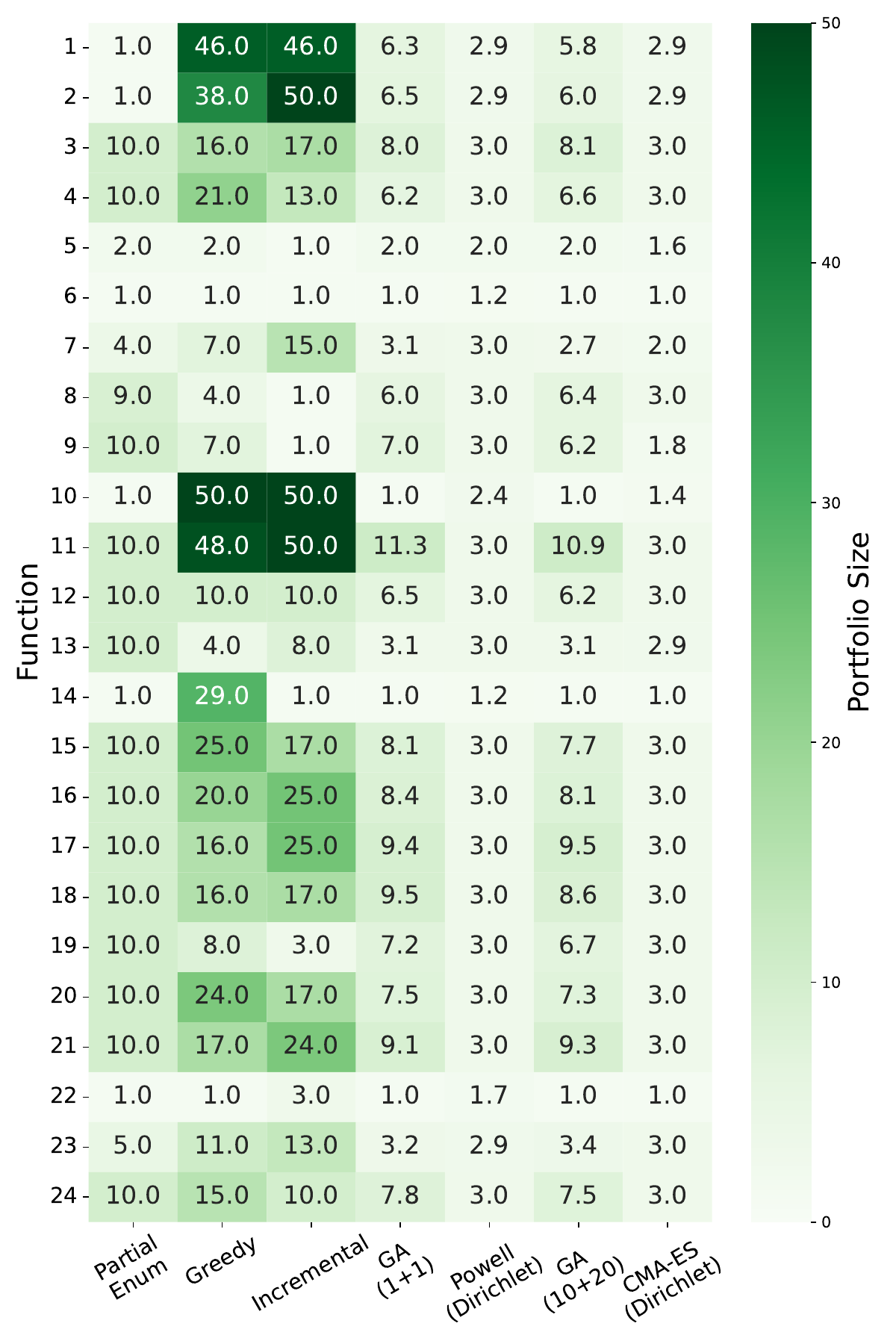}
        \caption{Average portfolio size.}
        \label{fig:1a_1f_appendix_size}
    \end{subfigure}
    \caption{Portfolio performance and size comparison for the case $\size{\A} = 1$, $\size{\F} = 1$.}
    \label{fig:1a_1f_appendix}
\end{figure}

\new{As shown in Figure~\ref{fig:1a_1f_appendix}, for several of the considered functions, distributing the available budget across multiple shorter independent runs yields better performance than allocating the same budget to a single long run. This effect is not universal: if the budget assigned to an individual run becomes too small, the algorithm may not have sufficient time to make meaningful progress. Nevertheless, the results indicate that, for some algorithm-function combinations, independent restarts can provide an effective trade-off between allowing sufficient search progress within each run and reducing the risk of spending a large fraction of the budget on an unfavorable search trajectory.}

Figure~\ref{fig:1a_1f_appendix_improvement} reveals that the greedy method consistently outperforms other approaches, except for function $F_{14}$, where it performs notably worse. Figure~\ref{fig:1a_1f_appendix_size} further shows that greedy portfolios tend to have larger sizes, reflecting a preference for multiple short runs rather than fewer long ones.

In contrast, black-box search methods, such as integer-based Genetic Algorithms and continuous optimization with Dirichlet projections (e.g., CMA-ES, Powell), struggle to compete. While they avoid major failures like those seen for greedy on $F_{14}$ or incremental search on $F_{22}$, they tend to explore only a limited part of the search space. This is reflected in the smaller portfolio sizes they generate, often failing to identify longer or more diverse portfolios.

\subsection{Two Functions - Three Algorithms}
\begin{figure}[tbp]
    \centering
    \begin{subfigure}[b]{0.48\columnwidth}
        \centering
        \includegraphics[width=\textwidth]{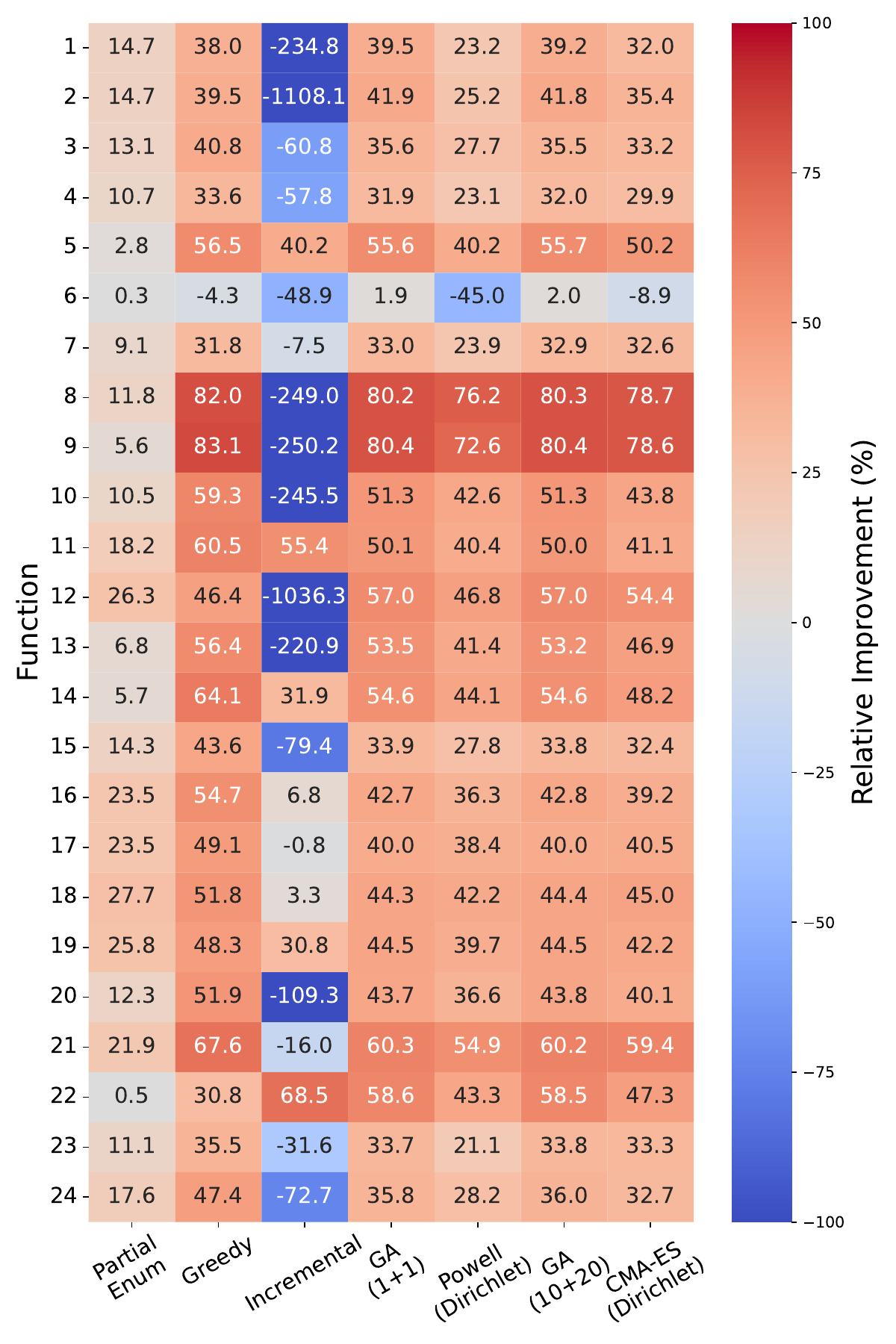}
        \caption{Relative Improvement (\%).}
        \label{fig:3a_2f_appendix_improvement}
    \end{subfigure}
    \hfill
    \begin{subfigure}[b]{0.48\columnwidth}
        \centering
        \includegraphics[width=\textwidth]{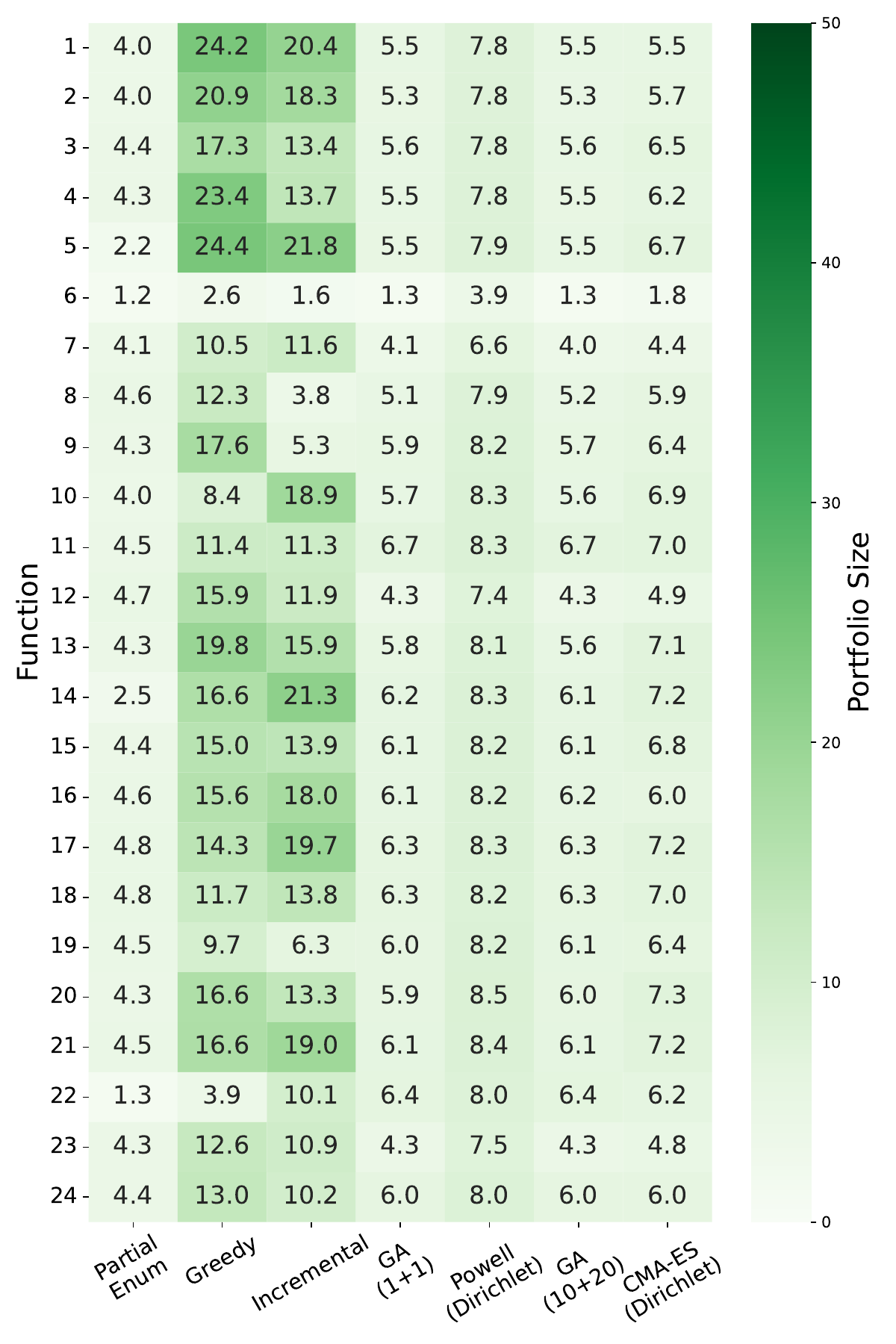}
        \caption{Average portfolio size.}
        \label{fig:3a_2f_appendix_size}
    \end{subfigure}
    \caption{Portfolio performance and size comparison for the case $\size{\A} = 3$, $\size{\F} = 2$.}
    \label{fig:3a_2f_appendix}
\end{figure}

The second experiment explores a more complex scenario, involving three algorithms ($\size{\A} = 3$) and two functions ($\size{\F} = 2$). In Figure~\ref{fig:3a_2f_appendix_improvement}, we observe how various methods perform under this increased complexity.
One immediate observation is that the incremental method struggles even more to find competitive portfolios in this setting. In contrast, black-box methods (e.g., Genetic Algorithms and continuous optimization with Dirichlet projections) become more competitive. This is also reflected in the portfolio sizes shown in Figure~\ref{fig:3a_2f_appendix_size}, where the greedy method no longer dominates in portfolio length, the sizes are now much closer across methods.

Importantly, we also observe a greater overall performance improvement compared to the single-function case. This suggests that algorithm complementarity starts to play a stronger role: different algorithms specialize in different functions, and the ability to exploit this diversity leads to more effective portfolios.

\subsection{All 24 Functions - One/Three Algorithms}
\begin{figure}[tbp]
    \centering
    \begin{subfigure}[b]{\columnwidth}
        \centering
        \includegraphics[width=\textwidth]{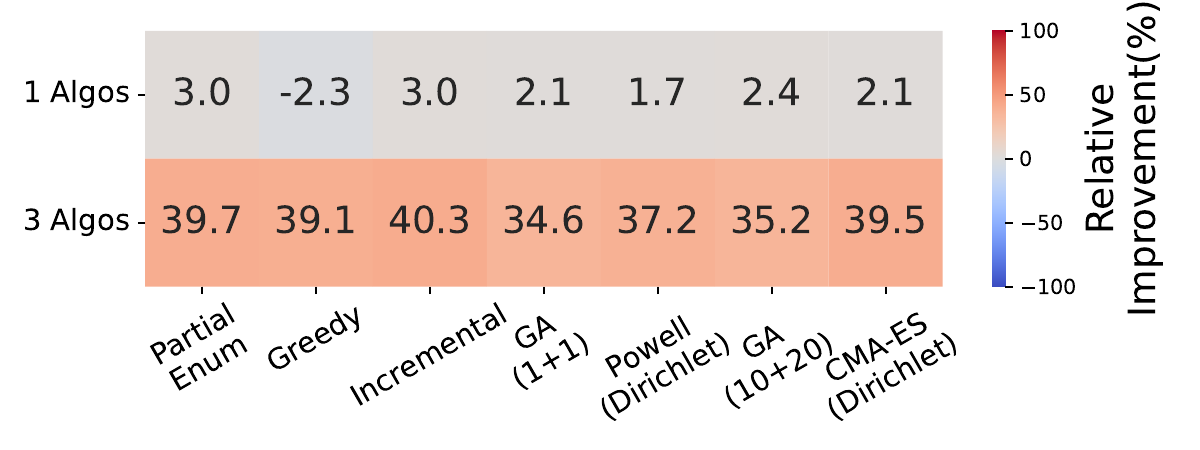}
        \caption{Relative Improvement (\%)}
        \label{fig:3a_24f_improvement}
    \end{subfigure}
    \vfill
    \begin{subfigure}[b]{\columnwidth}
        \centering
        \includegraphics[width=\textwidth]{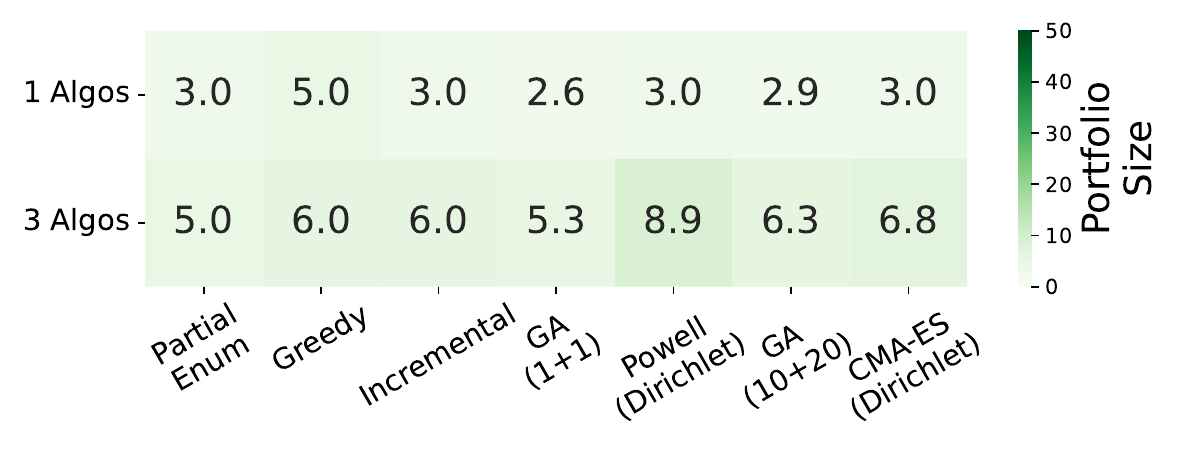}
        \caption{Average Portfolio Size}
        \label{fig:3a_24f_size}
    \end{subfigure}
    \caption{Portfolio performance and size comparison for the case $\size{\F} = 24$.}
    \label{fig:ternary_24f}
\end{figure}
For the final set of experiments, we consider the full set of 24 BBOB functions and compare two scenarios: using a single algorithm ($\size{\A} = 1$) versus using three algorithms ($\size{\A} = 3$). The main observation is that, when limited to a single algorithm, all methods struggle to construct high-performing portfolios, the diversity of functions makes it difficult to generalize effectively.

However, when multiple algorithms are available, the interplay between them enables significantly better performance, as different algorithms can complement each other across the diverse set of functions. In this case, we see that black-box methods approach the performance of the greedy strategy, a trend also reflected in the similarity in portfolio sizes across methods.

Even though the average portfolio size decreases as more functions are added, we still observe that splitting the budget across multiple independent runs within each algorithm remains beneficial for achieving strong performance.

\section{Equal Budget Splitting}
\label{app:equal-splits}

\begin{figure}[t]
    \centering
    \includegraphics[width=0.8\linewidth]
    {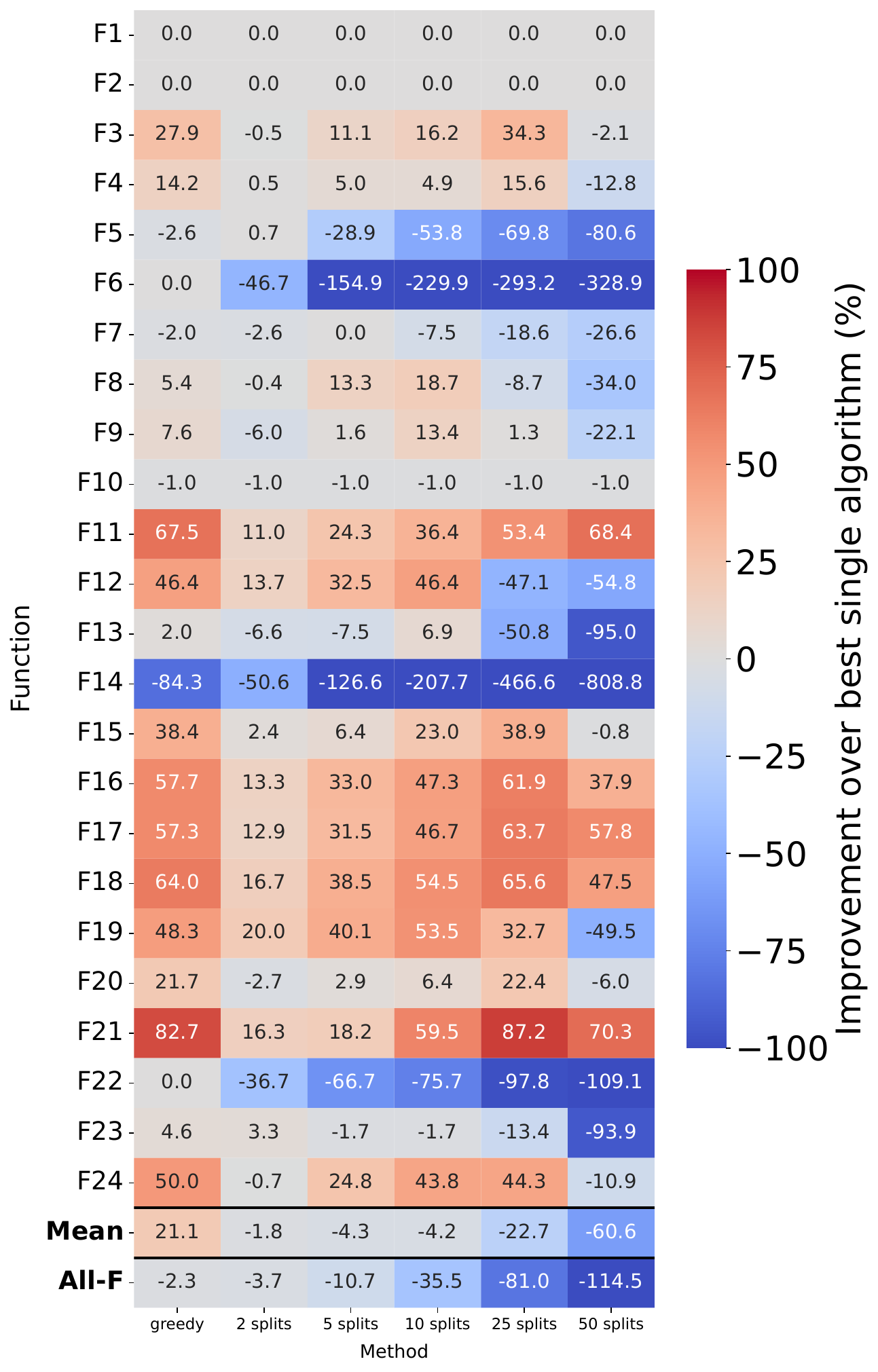}
    \caption{
    Relative improvement over the best single algorithm at a total budget of \(10{,}000\) function evaluations. The greedy method is compared with independently restarting a single algorithm using \(2\), \(5\), \(10\), \(25\), or \(50\) equal budget allocations. Rows \(F_1\)-\(F_{24}\) show function-specific results, \emph{Mean} reports the average over the function-specific constructions, and \emph{All-F} reports the portfolio constructed jointly over all functions. Positive values indicate improvement over the best single algorithm, while negative values indicate worse performance.}
    \label{fig:equal-splits}
\end{figure}

\new{A simple alternative to portfolio construction is to divide the available budget equally among several independent runs of a single algorithm. This strategy can benefit from restarts, but it does not optimize the number of runs or the budget assigned to each run. We therefore compare our greedy construction method with equal splitting into \(2\), \(5\), \(10\), \(25\), and \(50\) independent runs, using the same total budget of \(10{,}000\) function evaluations.}

\new{Equal splitting also becomes less practical when multiple algorithms are available. In that setting, the user must decide which algorithms to include, how much of the total budget each algorithm should receive, and how that allocation should be divided among repeated runs. These choices introduce additional decision variables and require prior knowledge about the behavior of the candidate algorithms. The proposed method instead selects the algorithm-budget pairs and their multiplicities as part of the portfolio-construction procedure.}

\new{Figure~\ref{fig:equal-splits} reports the relative improvement over the corresponding best single algorithm. Although some split configurations improve performance on individual functions, equal splitting is not consistently beneficial and generally deteriorates when the budget is divided among many short runs. In contrast, the greedy method achieves an average improvement of \(21.1\%\) across the function-specific constructions, compared with negative average improvements for all tested equal-split baselines.}

\new{These results show that the observed benefit cannot be explained solely by restarting an algorithm several times. Choosing the algorithms, the number of repetitions, and the budget assigned to each run are important parts of the portfolio-construction problem.}

\end{document}